\documentclass[11pt]{article}

\usepackage[preprint]{acl}
\usepackage{times}
\usepackage{latexsym}
\usepackage[T1]{fontenc}
\usepackage[utf8]{inputenc}
\usepackage{microtype}
\usepackage{inconsolata}
\usepackage{graphicx}
\usepackage{amsmath}
\usepackage{amssymb}
\usepackage{booktabs}
\usepackage{algorithm}
\usepackage{algpseudocode}
\usepackage{multirow}
\usepackage{float}
\usepackage{xcolor}
\usepackage{siunitx}

\title{Factorized Hypothesis Search for Evidence-to-Taxonomy Retrieval}

\author{
 \textbf{Linhai Ma\textsuperscript{1}},
 \textbf{Ethan F. Wei\textsuperscript{2}},
 \textbf{Xueqing Peng\textsuperscript{1}},
 \textbf{Yan Wang\textsuperscript{1,*}},\\
 \textbf{Lingfei Qian\textsuperscript{1,*}},
 \textbf{Víctor Gutiérrez-Basulto\textsuperscript{3}}
\\
\\
 \textsuperscript{1}The Fin AI, USA,
 \textsuperscript{2}Yale University, USA,
 \textsuperscript{3}Cardiff University, UK
\\
 \small{
   \textbf{Correspondence:} {\{wy2266336,lfqian94\}@gmail.com}
 }
}

\begin{document}
\maketitle

\begin{abstract}
Large-taxonomy retrieval often assumes that the input already expresses the target concept. In many settings, however, the input is indirect evidence, such as a table cell whose meaning depends on its row, column, datatype, and context. We call this mismatch the \emph{retrieval readiness gap}. Our analysis shows that the current index retrieves the target reliably when its semantics are explicit, while raw evidence often leaves it deep in the ranking. We propose Factorized Hypothesis Search (FHS), which maintains multiple partial interpretations over named semantic dimensions. These hypotheses support structured query rendering, multi-hypothesis retrieval, and dimension-level candidate verification. On both financial taxonomy tagging and CodiEsp clinical coding tasks, FHS achieves the best Recall@1, MRR, and final accuracy among the non-oracle methods. Replacing the factorized hypothesis path with a free-text ensemble causes the largest drop in head-ranking performance, while sequential refinement provides no additional gain over FHS’s strong parallel first round.
The code is available at \url{https://github.com/SarielMa/FHS}

% Large-taxonomy retrieval usually assumes the input already names the concept to retrieve. In many settings the input is instead indirect evidence, such as a table cell whose meaning depends on its row labels, column, and context. Query rewriting is not directly sufficient: serializing the evidence into one query still commits to a single reading. We call this the \emph{retrieval readiness gap}, and we find that what closes it is how the reading is represented, not how hard the system searches. A query-form comparison shows the gap is mainly a \emph{precision} problem: raw evidence reaches the right neighborhood but ranks the gold concept poorly, so grounding should adjudicate among readings rather than search deeper. We propose Factorized Hypothesis Search (FHS), which keeps several \emph{factorized} hypotheses instead of one rewrite. Because each hypothesis is a partial assignment over named dimensions, competing readings disagree on identifiable dimensions, so retrieval consensus can score them and a candidate-level verifier can reorder the head. Replacing the factorized hypotheses with a free-text ensemble causes the largest drop in Recall@1 and MRR among the FHS-specific ablations, while the definition-form rendering contributes more to final accuracy. Sequential refinement provides no additional gain over the strong parallel formulation.
\end{abstract}

\section{Introduction}

\begin{figure*}[htbp]
 \centering
 \includegraphics[width=0.8\textwidth]{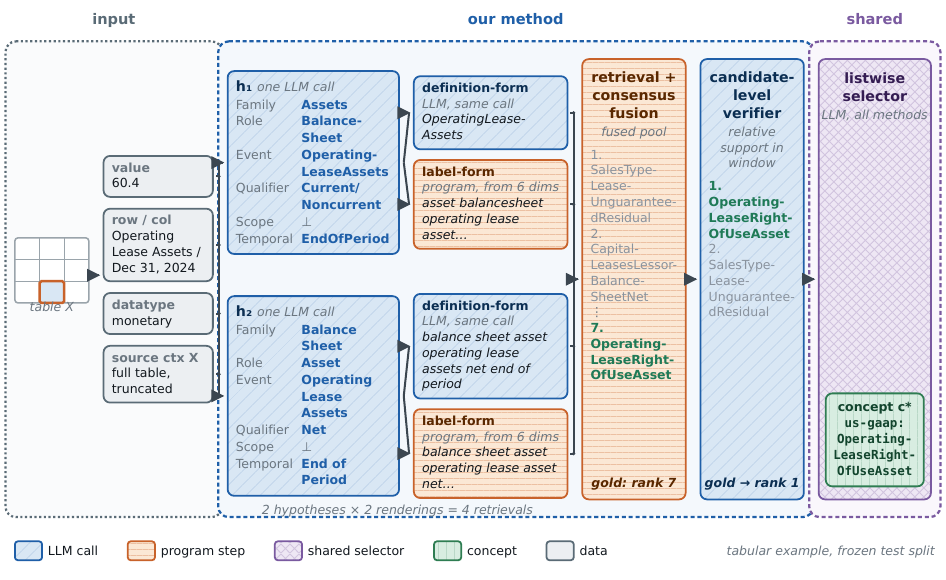}
 \caption{Overview of FHS on a tabular fact from the test split. FHS generates multiple factorized hypotheses from the target value and its context, renders each hypothesis into a definition-form query and a programmatically constructed label-form query, and fuses the retrieved rankings into a candidate pool. A candidate-level verifier then evaluates the candidates against the semantic commitments of the hypotheses. In this example, it moves the gold concept from rank 7 to rank 1. All compared methods share the final listwise selector.}

 \label{fig:main}
\end{figure*}
Many core NLP tasks, such as biomedical entity linking \citep{miranda2020overview,ye-mitchell-2025-llm}, schema matching \citep{semtab-2024,wang-etal-2025-linkalign}, and financial taxonomy tagging \citep{wang-etal-2025-fintagging}, require mapping an observed input to an entry in a large concept inventory. Standard approaches typically follow a retrieve-and-rerank paradigm \citep{wang-etal-2025-linkalign,ye-mitchell-2025-llm}, which succeeds when the input directly mentions or expresses the target concept. In practice, however, inputs often provide only indirect, contextual evidence. For example, interpreting a cell in a financial table requires synthesizing its numerical value, row and column headers, data type, and adjacent cells \citep{wang-etal-2025-fintagging}; similarly, assigning a code to a clinical result depends on the analyte, specimen type, and ambient clinical notes \citep{huang-etal-2022-plmicd}. In such settings, the target fact does not explicitly name its concept, while the surrounding context contains distracting cues pointing to unrelated entities. 
We term this fundamental structural discrepancy the \textbf{retrieval readiness gap} and formalize the setting of \emph{evidence-to-taxonomy retrieval}, where a model must synthesize a retrieval-ready representation of an isolated fact prior to searching a large taxonomy. This setup extends beyond the traditional query-document vocabulary mismatch addressed by hypothetical document embeddings (HyDE; \citealp{gao-etal-2023-precise}): because no explicit query exists, the system must infer the intended semantic concept directly from unstructured, noisy evidence.

A natural baseline is to rewrite the context into a single query \citep{wang-etal-2023-query2doc,gao-etal-2023-precise}. However, query rewriting inherently assumes that the underlying semantics are unambiguous and merely require lexical refinement. In our setting, the core semantics themselves are uncertain: an observed fact often admits multiple plausible interpretations, and prematurely committing to a single reading risks steering retrieval toward an incorrect sub-hierarchy of the taxonomy.
Our oracle probe (Section~\ref{sec:precision}, \ref{sec:rrg}) shows that the retriever
recovers the target reliably once it is specified by its label and
definition, whereas raw evidence ranks it deep in the candidate list.
The central bottleneck is therefore not candidate recall, but
constructing and scoring multiple interpretations so that the target
concept reaches the head of the ranking.

To address this challenge, we introduce \textbf{Factorized Hypothesis Search} (FHS). As illustrated in Figure~\ref{fig:main}, FHS maintains a set of explicit hypotheses regarding the underlying semantics of the target fact. Each hypothesis consists of a partial assignment over structured semantic dimensions, such as concept family, event type, qualifier, scope, and temporal context, leaving unsubstantiated dimensions explicitly unresolved. From a single generation call, each hypothesis derives both a definition-style query and a structured label-style query constructed from its resolved dimensions.  Retrieved rankings across hypotheses are consolidated into a unified candidate pool, which a candidate-level verifier then evaluates by checking each candidate against the explicit commitments of each hypothesis on a per-dimension basis. Factorization thus yields a shared schema for both query generation and candidate verification; contrasting with unstructured, free-text query sampling, where semantic commitments remain implicit.

Across financial tagging and clinical diagnosis coding, FHS improves head ranking and final prediction accuracy over direct retrieval and strong grounding baselines (Tables~\ref{tab:main_results}
and~\ref{tab:codiesp_main_results}), which suggests that FHS is not a
domain-specific solution. \textbf{Our contributions are as follows:}
\textbf{(1)} We formulate \emph{evidence-to-taxonomy retrieval}, where a located fact must be interpreted within its source context before it can serve as a retrieval query, and show by oracle analysis that query construction rather than index coverage is the binding constraint.
\textbf{(2)} We propose FHS, which represents plausible readings as factorized semantic hypotheses and uses the same representation for query rendering, multi-hypothesis retrieval, and dimension-level candidate verification.
\textbf{(3)} We instantiate FHS on both financial taxonomy tagging and
CodiEsp clinical coding. FHS achieves the best Recall@1, MRR, and final
accuracy among the compared non-oracle methods in both domains. Component
analyses identify the factorized hypothesis path (hypotheses + verifier) and definition-form rendering as the main sources of improvement. A
controlled study shows that iterative refinement does not justify its additional cost after a strong parallel round.

\section{Related Work}

\smallskip \noindent \textbf{Concept Inventory Alignment.}
Mapping observed mentions to entries of a structured inventory recurs across domains: biomedical entity linking~\citep{ye-mitchell-2025-llm,wang-etal-2025-aelc}, medical coding over tens of thousands of codes~\citep{huang-etal-2022-plmicd}, schema linking~\citep{wang-etal-2025-linkalign}, table cell linking with row and column context~\citep{wang-etal-2024-rocel,zhou-etal-2024-gendecider,semtab-2024}, and XBRL tagging, where linking accuracy stays low even for strong LLMs on tables~\citep{wang-etal-2025-fintagging}. The shared difficulty is a dense inventory of near-neighbors and an input that must be interpreted rather than matched; we abstract it as evidence-to-taxonomy retrieval and address the grounding stage before retrieval. Supervised closed-set methods~\citep{huang-etal-2022-plmicd,wang-etal-2024-rocel,zhou-etal-2024-gendecider} need task-specific training over a fixed label set, so they are cross-domain motivation rather than zero-shot comparators.

\smallskip \noindent \textbf{Context-Aware Entity Linking.}
Conventional context-aware entity linking generates candidates for a
mention and ranks them with the mention encoded together with its
context~\citep{sevgili-etal-2022-neural}, or enriches the mention query
with LLM-generated feedback~\citep{shlyk-hunter-2026-improving}. In
both cases the mention already refers to the
target concept and serves as the query. The difference in our setting
is the input. A table cell value rarely names its concept, and the
shared context does not resolve it, because one table hosts
many facts pointing to different concepts (Section~\ref{sec:evi_to_tax}).
The query is therefore not given but constructed. FHS keeps several
factorized readings of it rather than a single context encoding.

\smallskip \noindent \textbf{Query Transformation for Retrieval.}
HyDE~\citep{gao-etal-2023-precise} and
Query2Doc~\citep{wang-etal-2023-query2doc} expand an explicit query
into a pseudo-document, addressing a query--document representation
gap. A second line of work rewrites incomplete or context-dependent
queries~\citep{li-etal-2025-dialogue,qin-etal-2025-reinforced,
ye-etal-2025-q}, while a third enriches the document
side~\citep{liang-etal-2025-improving}. Our input is not a query to
reformulate, but a located fact whose intended concept must be inferred
from its source context. We adapt HyDE and Query2Doc as the basis of
our one-pass free-text grounding baseline. In contrast, FHS maintains
multiple competing interpretations rather than committing to a
single rewrite.

% \smallskip \noindent \textbf{Iterative and Multi-step Retrieval.}
% Prior approaches alternate clarification or rewriting with retrieval~\citep{cao-etal-2025-icr,wang-etal-2025-aelc,fang-glass-2026-beyond}; Self-RAG~\citep{asai-etal-2024-selfrag} and ITER-RETGEN~\citep{shao-etal-2023-iterretgen} interleave retrieval with generation or self-critique, and others align rewriters with retriever preferences~\citep{yoon-etal-2025-ask,cao-etal-2026-multi}. These methods assume an initial explicit query. Our FHS utilizes this idea: iteration helps when the initial representation is weak and stops being useful once a parallel ensemble is near its ceiling (Section~\ref{sec:negative}). The retrieve-revise loop methods thus serve as our retrieval-feedback baselines. Trained-reflection and clarification variants need retraining or \textcolor{red}{a user turn this task lacks}\nb{V: I have no idea what this means. Do we need it?}, so they are not selected as methods for comparison.

\smallskip \noindent \textbf{Iterative and Multi-step Retrieval.}
Prior approaches alternate query clarification or rewriting with
retrieval~\citep{cao-etal-2025-icr,wang-etal-2025-aelc,
fang-glass-2026-beyond}. Self-RAG~\citep{asai-etal-2024-selfrag}
and ITER-RETGEN~\citep{shao-etal-2023-iterretgen} interleave
retrieval with generation or self-critique, while other methods align
query rewriters with retriever preferences~\citep{yoon-etal-2025-ask,
cao-etal-2026-multi}. These methods generally assume that an explicit initial query is
available. We adapt them to our setting through intrinsic
self-refinement and retrieval-feedback refinement baselines, and
introduce FHS-Seq as a controlled sequential counterpart to FHS.
Iteration improves over weak single-pass grounding but not over FHS's
strong parallel first round (Section~\ref{sec:negative}). Methods
requiring retraining or interactive user clarification fall outside
our zero-shot, offline evaluation setting.

\paragraph{Self-Correction, Verification, and Aggregation.}
Intrinsic self-correction can make models waver~\citep{zhang-etal-2025-understanding} and confidence trades off against critique~\citep{yang-etal-2025-confidence}, which motivates verification~\citep{song-etal-2025-progco}. We use the LLM not as an absolute self-verifier but as a \emph{relative} verifier over retrieved candidates. Self-consistency resolves independent samples by majority vote~\citep{wang-etal-2023-selfconsistency}; our ensemble is also sampled, but factorization makes the samples disagree on identifiable dimensions. The retrievals are fused by cross-hypothesis consensus instead of voting over whole outputs (Table~\ref{tab:ablation} isolates the ensemble from the aggregation rule).

\section{Problem Formulation}

\subsection{Evidence-to-Taxonomy Retrieval}
\label{sec:evi_to_tax}

Let $\mathcal{T} = \{c_1,\ldots,c_N\}$ be a taxonomy, where each $c_j$ is associated with a canonical label, definition, and structural metadata. The input is not a standalone query but a \emph{located fact within a shared source context}. A \emph{source context $X$} is a single table or narrative passage and typically contains many facts (e.g., 21.3 per table on average in the financial tagging data, cf.\ Table~\ref{tab:data_stats}), each grounding to a different concept.
A fact $x=(\ell,a,X)$ identifies one target within its source context,
where $\ell$ denotes the target locus and $a$ its observed content.
For tabular evidence, $\ell$ specifies the cell location and datatype,
and $a$ is the cell value; for narrative evidence, $\ell$ is the
mention span and $a$ is its surface text. The goal of
evidence-to-taxonomy retrieval is to identify, for each fact
independently, the concept $c^*\in\mathcal T$ that matches the located
cell or mention.
% A \emph{fact $x=(\ell,a,X)$} singles out one target: a cell with locus identifier $\ell$ (e.g., its location and datatype) and value $a$ in table $X$, or an entity mention at span $\ell$ in passage $X$\nb{V: In this second case when we have a passage, what is $a$?}. The goal\nb{the goal of what? Maybe mention explicitly the name of the task.} is to identify, \emph{for each fact independently}, which concept $c^*\in\mathcal T$ matches the located cell or mention.
We note two properties separating this from conventional retrieval. First, the context $X$ is not a query: the same $X$ hosts many facts pointing to different concepts, so it under-determines any single target. Second, the locus $(\ell,a)$ is not a query either, e.g., a bare numeric cell or a short mention rarely names its concept. A query must therefore be \emph{constructed} from the located fact by reading $X$. Let $\mathcal Q$ denote the space of retrieval queries. For a fixed taxonomy $\mathcal T$, a retriever $R_{\mathcal T}:\mathcal Q\rightarrow\mathcal T^K$ maps a query to an ordered list of $K$ candidate concepts. In conventional retrieval, the input is already a query in
$\mathcal Q$. In our setting, the input is instead a located fact
$x=(\ell,a,X)\in\mathcal F$, where $\mathcal F$ denotes the space of
located facts; thus, $x\notin\mathcal Q$. A grounding function
$g:\mathcal F\rightarrow\mathcal Q$ must therefore construct
$q=g(x)$, a retrieval-ready query expressing the target concept.
% A \emph{retriever $R:\mathcal Q\times\mathcal T\to\mathcal T^K$}\nb{V: What is $\mathcal Q$? It is is undefined. Is it the query. If so, the mapping maps a pair of a query and an elements of the taxonomy to ...} maps a query to $K$ ordered candidates. 

% In conventional retrieval the input is itself the query. In this case, the fact $x\notin\mathcal Q$\nb{$x$ seems to be special. What is it?}, so a \emph{grounding function} $g:\mathcal F\to\mathcal Q$, on the space $\mathcal F$ of located facts, must first produce $q=g(x)$, enriching the fact into a query that expresses the target concept.

\subsection{The Gap Is Interpretive}
\label{sec:rrg}

% Let $g^*$ denote an oracle grounding\nb{Is an oracle grounding the same as a grounding? What is an oracle?}. The retrieval readiness gap is
% \begin{equation}
% \operatorname{sim}\!\left(\phi(x),\phi(c^*)\right)
% \;\ll\;
% \operatorname{sim}\!\left(\phi(g^*(x)),\phi(c^*)\right),
% \label{eq:gap}
% \end{equation}
% where $\phi$ is the retriever's representation and $\operatorname{sim}$ is a similarity function \nb{Why not the standard $<$ symbol?}. 
Let $g^*:\mathcal F\rightarrow\mathcal Q$ denote an oracle grounding
function that constructs a query explicitly expressing the gold
concept, and let $q_{\mathrm{raw}}(x)$ denote the direct serialization
of the located fact used for retrieval. We characterize the retrieval
readiness gap as
\begin{equation}
\scriptsize
\Delta(x)=
\operatorname{sim}\!\left(\phi(g^*(x)),\phi(c^*)\right)
-
\operatorname{sim}\!\left(\phi(q_{\mathrm{raw}}(x)),\phi(c^*)\right)
>0,
\label{eq:gap}
\end{equation}
where $\phi$ is the retriever's representation and
$\operatorname{sim}$ is its similarity function.
Unlike the query--document distribution mismatch addressed by hypothetical document generation~\citep{gao-etal-2023-precise} and query expansion~\citep{wang-etal-2023-query2doc}, the gap here arises from \emph{interpretive ambiguity}: the fact admits multiple plausible readings, and the system must determine which one is correct. Equation~\eqref{eq:gap} provides a conceptual characterization of the gap. We assess its retrieval-level implication using an oracle-query probe. Substituting each gold concept's own canonical label and definition for $g^*(x)$, returns the target within the top ten for every test fact (Appendix~\ref{app:probe}). 
Given a query that expresses the target, the index already reaches it,
so the missing component is the query rather than index capacity. The
index still matters: its label-coverage term helps every method that
uses it (Table~\ref{tab:ablation}), but it is shared across methods.
% This result suggests that index capacity is not the primary bottleneck in
% this setting. This does not imply that the index is unimportant; our index carries a label-coverage term, and removing it costs recall for every method that uses the index (Table~\ref{tab:ablation}). The probe captures something more specific: given a query that expresses the target, the index already reaches it, so the missing component must be the query.

\subsection{The Gap Is Primarily About Precision}
\label{sec:precision}

\begin{table}[t]
\centering
\scriptsize
\setlength{\tabcolsep}{3pt}
\begin{tabular}{@{}p{2.6cm}cccc@{}}
\toprule
\textbf{Query representation} & \textbf{R@10} & \textbf{R@50} & \textbf{R@200} & \textbf{MRR} \\
\midrule
Gold concept label + definition (probe) & 1.000 & 1.000 & 1.000 & 0.972 \\
Raw context (direct retrieval) & 0.240 & \textbf{0.562} & \textbf{0.752} & 0.111 \\
Free-text grounding & 0.304 & 0.496 & 0.734 & 0.171 \\
Structured, def-form-only & \textbf{0.353} & 0.521 & 0.657 & \textbf{0.195} \\
\bottomrule
\end{tabular}
\caption{How far four query representations get. The probe queries with the gold concept's own label and definition and bounds what the index can do.}
\label{tab:probe}
\end{table}

% A second measurement constrains what a grounding method should optimize. \nb{V: I couldnt make sense of the following sentence}\textcolor{red}{The located fact serialized directly as a query places $c^*$ within the top 200 on 75.2\% of facts but within the top 10 on only 24.0\%.} 

A second measurement clarifies what a grounding method should
optimize. When the raw located fact is serialized and used directly
as the retrieval query, the gold concept appears in the top 200 for
75.2\% of facts but in the top 10 for only 24.0\%. Thus, the gold
concept is often retrieved but usually ranked far from the head of
the list.
The target is usually reachable but  badly placed. The dominant bottleneck is thus about position rather than about reachability. Grounding acts on position: a single structured hypothesis raises top-10 recall to 35.3\% and MRR (Mean Reciprocal Rank) from 0.111 to 0.195, while top-200 recall falls to 65.7\% (Table \ref{tab:probe}). The accumulated pool over $J$ hypotheses recovers part of the lost depth. 
% \nb{V: I modified the next sentence to what I believed we wanted to say - the previous version was gramatically incorrect}
For verification, we design a score measuring how well competing interpretations align with the returned candidates, rather than measuring coverage expansion. Table~\ref{tab:main_results} shows the same trade-off across the compared methods: FHS performs best at the head, whereas iterative methods recover more gold concepts at deeper cutoffs. What distinguishes the methods is therefore not a uniform improvement across ranks, but how they trade head ranking against retrieval depth. The gap also accounts for most of the remaining difficulty. 
On Financial Tagging, raw evidence reaches Recall@1 of 0.041 and FHS 0.185 against 0.955 for the gold-query probe (Table~\ref{tab:main_results}), closing 16\% of that distance. On CodiEsp, direct retrieval gives Recall@1 of 0.046 and FHS 0.264 against an oracle of 0.998 (Table~\ref{tab:codiesp_main_results}), closing 23\%. About four fifths of the gap remains open on both tasks, which is what keeps the absolute scores low.
% On Financial Tagging, raw evidence reaches top-10 recall of 0.240 and the
% gold-query probe 1.000 (Table~\ref{tab:probe}). FHS reaches 0.397
% (Table~\ref{tab:main_results}), closing 21\% of that distance. On
% CodiEsp, direct retrieval gives Recall@1 of 0.046 and FHS 0.264 against
% an oracle of 0.998 (Table~\ref{tab:codiesp_main_results}), closing
% 23\%. About four fifths of the gap remains open on both tasks, which
% is what keeps the absolute scores low.

\subsection{Factorized Interpretations}
\label{sec:factorized}

The interpretive ambiguity has structure. Each $c\in\mathcal T$ can be characterized by semantic attributes along $M$ dimensions.
We use $\mathcal D$ to denote that dimension set and $\mathcal V$ to denote the values its dimensions may take. A dimension is \emph{resolved} when a hypothesis assigns it a specific value in $\mathcal V$; otherwise, it remains unresolved. The label-style query is built from the resolved dimension values, and the candidate-level verifier of Section~\ref{sec:llmverifier} returns a verdict per dimension.
A fact may give strong signal on some dimensions, such as a column header indicating a monetary measurement, while leaving others latent, such as whether the value is net of adjustments or which period it covers. A \emph{factorized semantic hypothesis} is a partial assignment $h:\mathcal D\rightarrow\mathcal V\cup\{\bot\}$, where $\bot$ denotes a set of unresolved dimensions. %Unlike a free-form rewrite, this exposes which commitments have been made and which remain open, which is what makes dimension-level verification possible at all, and what allows an abstention to be treated differently from a wrong answer.
Unlike free-form rewrites, this explicitly reveals which semantic commitments are resolved and which remain open; providing the foundation for dimension-level verification and enabling the model to distinguish intentional abstentions from incorrect answers.

\subsection{Ensemble and Objective}
\label{sec:partial}
\label{sec:objective}

Different hypotheses render to different queries and retrieve overlapping but distinct candidate sets. Useful complementarity comes only from hypotheses that resolve an open dimension differently yet plausibly, not from forcing hypotheses apart. In a controlled pilot, explicitly prompting the hypotheses to be
diverse reduced both their pairwise overlap and their accumulated
coverage. Although the generated hypotheses were more distinct,
their individual retrieval quality decreased by 3--8 Recall@200
points (Appendix~\ref{app:pilot}). We therefore sample an unmodified generator and let the verifier and aggregator exploit whatever complementarity arises. Given a fact, the system produces $J$ hypotheses, each yielding one
or more retrieval rankings. Let $r$ index these rankings and
$\mathcal C_r$ denote the top-$K$ candidate list returned by ranking
$r$. The system pools their candidates,
$\mathcal U=\bigcup_r\mathcal C_r$, and consolidates them into a
fixed-size list $\widehat{\mathcal C}_K$, with objective
$\max\Pr(c^*\in\widehat{\mathcal C}_K)$. We also report search coverage $\Pr(c^*\in\mathcal U)$ as a diagnostic. Coverage bounds the objective but it is not the goal, since a system can reach high coverage and still fail to consolidate. The consolidation loss is 7.4 points for the selected stochastic-sampling configuration on the development data (Appendix~\ref{app:pilot}).

\section{FHS: Factorized Hypothesis Search}
\label{sec:method}

%\subsection{Overview}

%FHS runs four steps over $J$ hypotheses, detailed below: \emph{generation, rendering, consensus fusion}, and \emph{candidate reranking}; Algorithm~\ref{alg:ags} in Appendix~\ref{app:floats} provides the pseudocode. The $J$ generation and  reranking calls are parallelizable; rendering, retrieval, and fusion are program-driven. The retriever, the index and its coverage term, and the shared listwise selector are standard components described with the setup (Section~\ref{sec:fintag}).

FHS operates across four main stages over $J$ hypotheses: \emph{generation}, \emph{rendering}, \emph{consensus fusion}, and \emph{candidate reranking} (detailed in Algorithm~\ref{alg:ags}, Appendix~\ref{app:floats}). The $J$ generation and reranking calls are fully parallelizable, whereas rendering, retrieval, and fusion are executed programmatically. The underlying retriever, the search index and its associated coverage term, and the shared listwise selector are standard components defined in Section~\ref{sec:fintag}.

\subsection{Factorized Hypotheses}
\label{sec:generators}

The generator emits $J$ hypotheses under stochastic decoding. Each is a partial assignment over a fixed set of named dimensions (Section~\ref{sec:factorized}), with unsupported dimensions left blank rather than guessed. Blanks are cheap:
% \nb{V: Nice explanation :), I was wondering why not to simply guess }
 an unresolved dimension is skipped by the renderer and excluded from later scoring, so caution costs nothing. Guessing is not free: a wrong value enters the query and pulls retrieval toward a region the fact does not support.

% \smallskip \noindent \textbf{Why sampling rather than forced variation.}
% We sample an unmodified generator rather than forcing the hypotheses to differ. Section~\ref{sec:partial} shows why: prompting for diversity, or assigning dimensions per generator, lowers coverage because it degrades each hypothesis. Two senses of coverage are in play: Section~\ref{sec:precision}  highlights that the task is not primarily coverage-bound, because the gold concept is usually already inside the retrieved pool. %The present claim is about the \emph{pool} the ensemble accumulates, which forced diversity shrinks. Neither claim implies the other.
% The present claim concerns the candidate \emph{pool} accumulated by the ensemble, which enforced diversity inadvertently constrains; neither claim implies the other.

\smallskip \noindent \textbf{Why sampling rather than forced variation.}
We sample an unmodified generator rather than forcing the hypotheses to
differ, because prompting for diversity degrades each hypothesis and
shrinks the candidate \emph{pool} the ensemble accumulates
(Section~\ref{sec:partial}). This is a different sense of coverage from
Section~\ref{sec:precision}, where the gold concept is usually already
inside the pool. Neither claim implies the other.

\smallskip \noindent \textbf{What factorization buys.}
Sampling $J$ candidate outputs shares the initial generation mechanism of self-consistency \citep{wang-etal-2023-selfconsistency}; the core distinction lies in how downstream fusion utilizes these outputs. Because our hypotheses are factorized, disagreements across samples map onto identifiable semantic dimensions rather than competing opaque strings, allowing each hypothesis to render into a distinct yet semantically plausible query. Table~\ref{tab:ablation} isolates these effects by separately evaluating two baselines: reducing the ensemble to a single hypothesis and replacing factorized hypotheses with a free-text ensemble. 

\subsection{Query Rendering}
\label{sec:render}

Each hypothesis is issued as a pair of queries $\bigl(q^{\mathrm{lab}},\,q^{\mathrm{def}}\bigr)$, where one is written by an LLM and the other by rules. %The label-style $q^{\mathrm{lab}}$ concatenates the resolved dimension values in a fixed order, tokenized exactly as the index tokenizes labels (Eq.~\eqref{eq:coverage}). 
The label-style query $q^{\mathrm{lab}}$ concatenates resolved dimension values in a canonical sequence, tokenized using the exact tokenisation scheme applied to index labels (Eq.~\eqref{eq:coverage}). It issues no query, if no dimension is resolved. As for the definition-style  $q^{\mathrm{def}}$: the generator emits it in the same structured call. Both are prefixed with the cell's own identifier. 
% \textcolor{red}{as is every other method's query}.\nb{V: I dont know what this means, or even if it is needed} 
So the identifier is always present and what differs between methods is only what is added to it. %Dual rendering follows from Section~\ref{sec:rrg}: a concept's own label retrieves it reliably, so a query that imitates label structure targets a lexical-form mismatch that composes with interpretive ambiguity. On narrative evidence, where the concept is usually stated in words, we use the definition form alone. In Appendix~\ref{app:render} we provide the tokenizer, the fallbacks and an example.
Dual rendering directly addresses the retrieval readiness gap defined in Section~\ref{sec:rrg}: because a concept's canonical label yields reliable retrieval, a query that mirrors label structure explicitly targets lexical-form mismatch alongside interpretive ambiguity. For narrative evidence, where concepts are typically described in natural prose, we rely exclusively on definition-style queries. Details regarding the tokenisation scheme, fallback rules, and concrete examples are provided in Appendix~\ref{app:render}.

\subsection{Consensus Fusion}
\label{sec:aggregator}

%Let $r$ index the rankings produced across all hypotheses and renderings, with pool $\mathcal U=\bigcup_r\mathcal C_r$. We fuse by summed reciprocal rank $S(c)$, the standard RRF form~\citep{cormack2009reciprocal} with the sum taken over the rankings that returned $c$ (Eq.~\ref{eq:sumrrf}, Appendix~\ref{app:fusion}). Table~\ref{tab:ablation} reports the substitution on test, where the two options are within uncertainty. We present summation as a selection rather than a component claim. Scores are then range-normalized over the pool to $\widetilde S(c)$, which together with the corresponding hypothesis serves as the input to the verifier below.
Let $r$ index the candidate rankings generated across all hypotheses and query renderings, yielding the unified candidate pool $\mathcal{U} = \bigcup_r \mathcal{C}_r$. We aggregate candidate scores using RRF~\citep{cormack2009reciprocal}, where $S(c)$ sums reciprocal ranks strictly over the subset of rankings containing concept $c$ (Eq.~\eqref{eq:sumrrf}, Appendix~\ref{app:fusion}). As reported in Table~\ref{tab:ablation}, comparing this variant against alternative fusion methods on the test set shows performance within margin of error; we thus present summation as a deliberate design choice rather than an isolated component claim. Finally, scores are range-normalised over $\mathcal{U}$ to yield $\widetilde{S}(c)$, which, along with its corresponding hypothesis, serves as input to the verifier below.

\subsection{Candidate-Level Verifier}
\label{sec:llmverifier}

For each hypothesis $h_j$, we construct a window $\mathcal B_j$ of
$K_v$ candidates from the fused ranking $\widetilde S$. Because its head
may contain many lexical near-duplicates, we scan the top 60 candidates
and retain the highest-ranked candidate from each distinct category
profile. This produces a window covering structurally different
interpretations. Fallback rules and window-size sensitivity are reported
in Appendix~\ref{app:window}. A structured LLM call receives the observed fact $x$, hypothesis $h_j$, and the candidates in $\mathcal B_j$. For each candidate and semantic dimension, it returns \textit{support}, \textit{no support}, or
\textit{abstain}. We define $v_j(c)\in[0,1]$ as the fraction of
non-abstaining verdicts that support candidate $c$. Candidates outside
$\mathcal B_j$ receive the mean support within the window:
\begin{equation}
\footnotesize
\widetilde v_j(c)=
\begin{cases}
v_j(c), & c\in\mathcal B_j,\\[2pt]
\dfrac{1}{|\mathcal B_j|}
\displaystyle\sum_{c'\in\mathcal B_j}v_j(c'),
& c\notin\mathcal B_j,
\end{cases}
\quad
\overline v(c)=\frac{1}{J}\sum_{j=1}^{J}\widetilde v_j(c)
\label{eq:llm_support}
\end{equation}
The verifier support is combined with the normalized retrieval score:
\begin{equation}
\footnotesize
S_{\mathrm{final}}(c)
=\widetilde S(c)+\beta\,\overline v(c),\qquad
\widehat{\mathcal C}_K
=\operatorname{TopK}_{c\in\mathcal U}S_{\mathrm{final}}(c).
\label{eq:hybrid_score}
\end{equation}
The verifier therefore only reorders candidates already present in the
retrieved pool. Additional scoring conventions are detailed in
Appendix~\ref{app:window}.

\subsection{Sequential Refinement: A Negative Control}
\label{sec:negative}

The natural agentic alternative to parallel hypothesis generation is
sequential refinement, in which retrieved candidates provide feedback for
iteratively revising the hypotheses. We implement \mbox{FHS-Seq} by replacing
FHS's one-shot control flow with a revise-and-refetch loop while keeping the
hypothesis representation, retrieval pipeline, and candidate-level verifier
fixed. Its first round is identical to the parallel round of FHS, allowing the
comparison between round one and the full episode to isolate the effect of
subsequent sequential refinement. 
The additional rounds change
the top-50 candidate pool but produce no significant improvement in Recall@50
(Table~\ref{tab:seq_outcome}) and yield lower final accuracy than FHS
(Table~\ref{tab:main_results}), despite the extra inference rounds. The oracle result in
Table~\ref{tab:ablation} helps explain this outcome: perfect selection among
FHS's existing hypotheses offers only limited additional headroom, particularly
for final accuracy. Appendix~\ref{app:seq} provides the oracle calculations and
round-level diagnostics.

\section{Experiments}
\label{sec:experiments}
To evaluate the effectiveness of FHS, we investigate the following research questions: 
\noindent \textbf{(RQ1)} Does FHS improve grounding accuracy compared to direct retrieval, single-pass, and hypothesis-count-matched parallel baselines? 
\textbf{(RQ2)} What are the individual contributions of factorized representations, ensembling, dual rendering, and consensus fusion? 
\textbf{(RQ3)} Does candidate-level verification provide gain beyond the fused ranking? 
\textbf{(RQ4)} Does sequential revision yield improvements over a single parallel pass?

\subsection{Task Instantiation}
\label{sec:fintag}

We evaluate on the financial tagging benchmark of
\citet{wang-etal-2025-fintagging} using the 2024 US-GAAP inventory of
17,388 concepts. Each instance contains a source context and an annotated
target locus, including its value and datatype. We therefore evaluate
grounding and retrieval given a located fact, without modeling upstream fact
extraction. We instantiate FHS using the six semantic dimensions in
Table~\ref{tab:schema}. Dataset statistics, context serialization, inventory
construction, and dimension-matching rules are provided in
Appendix~\ref{app:fintagging}.
% \paragraph{CodiEsp Diagnosis Coding.}
We additionally evaluate on the diagnosis portion of CodiEsp
\citep{miranda2020overview,miranda2020codiesp}, using an ICD-10-CM inventory
of 71,344 candidate codes. Each instance contains an English clinical context
and a relocated diagnosis mention. As in Financial Tagging, we evaluate
retrieval given the located mention rather than upstream mention extraction.
The corresponding six-dimensional instantiation is shown in
Table~\ref{tab:codiesp_schema}; dataset preparation, mention relocation,
inventory construction, and matching rules are detailed in
Appendix~\ref{app:codiesp}.

\begin{table}[t]
\centering
\scriptsize
\setlength{\tabcolsep}{2pt} % Reduces horizontal space between columns
\begin{tabular}{@{}llll@{}}
\toprule
\textbf{Dimension} & \textbf{Meaning} & \textbf{Example} & \textbf{Match} \\
\midrule
\textsc{Family} & broad accounting family & Asset & vocab (14) \\
\textsc{Role} & statement or semantic role & Balance-sheet item & overlap \\
\textsc{Event} & underlying transaction & Operating lease & overlap \\
\textsc{Qualifier} & measurement or modifier & Net & vocab (18) \\
\textsc{Scope} & dimensional context & Reportable segment & vocab (7) \\
\textsc{Temporal} & time interpretation & Duration, current year & vocab (11) \\
\bottomrule
\end{tabular}
\caption{The $M{=}6$ dimensions instantiated for 2024 US-GAAP. \textbf{Match} is
how the dimension is compared against a candidate. The four vocabulary-matched
dimensions form the category profile.}
\label{tab:schema}
\end{table}

 \begin{table}[t] 
\centering 
\scriptsize 
\setlength{\tabcolsep}{2pt} % Reduces horizontal padding (default is usually 6pt)
\begin{tabular}{@{}llll@{}} 
\toprule 
\textbf{Dimension} & \textbf{Meaning} & \textbf{Example} & \textbf{Match} \\ 
\midrule 
\textsc{Family} & broad clinical family & Respiratory diseases & vocab (21) \\ 
\textsc{Role} & diagnosis-code class & Injury/poisoning & vocab (9) \\ 
\textsc{Event} & specific condition or finding & Costal fracture & overlap \\ 
\textsc{Qualifier} & modifier or severity & Acute, malignant, open & vocab (21) \\ 
\textsc{Scope} & laterality & Left, bilateral & vocab (5) \\ 
\textsc{Temporal} & encounter or extension status & Initial encounter & vocab (12) \\ 
\bottomrule 
\end{tabular} 
\caption{The $M{=}6$ dimensions instantiated for CodiEsp diagnosis coding.} 
\label{tab:codiesp_schema} 
\end{table}

% \subsubsection{Financial Tagging}
% \label{sec:fintag}

% We instantiate evidence-to-taxonomy retrieval on the financial tagging dataset and the 2024 US-GAAP taxonomy. Each fact is $x_i=(X_i,\ell_i,v_i)$: source context $X_i$, a locus identifier $\ell_i$ (target location together with its datatype), and target value $v_i$. The full taxonomy contains 17,388 concepts before task-specific filtering. We instantiate the dimension schema of Section~\ref{sec:factorized} for US-GAAP with $M=6$ designed dimensions, drawn from the taxonomy's structure as is standard in schema-linking and clinical coding: \textsc{Family}, \textsc{Role}, \textsc{Event}, \textsc{Qualifier}, \textsc{Scope}, and \textsc{Temporal} (Table~\ref{tab:schema}). See Appendix~\ref{app:fintagging} for more details.

% \begin{table}[!htbp]
% \centering
% \scriptsize
% \begin{tabular}{@{}lll@{}}
% \toprule
% \textbf{Dimension} & \textbf{Meaning} & \textbf{Example} \\
% \midrule
% \textsc{Family} & broad accounting family & Asset \\
% \textsc{Role} & statement or semantic role & Balance-sheet item \\
% \textsc{Event} & underlying transaction & Operating lease \\
% \textsc{Qualifier} & measurement or modifier & Net \\
% \textsc{Scope} & dimensional context & Reportable segment \\
% \textsc{Temporal} & time interpretation & Duration, current year \\
% \bottomrule
% \end{tabular}
% \caption{The $M{=}6$ dimensions instantiated for 2024 US-GAAP. }
% \label{tab:schema}
% \end{table}

% \subsubsection{some setttings}

\subsection{Compared Methods}
\label{sec:baselines}

%\textbf{Single-pass and parallel:} direct retrieval~\citep{wang-etal-2025-fintagging}; one-pass grounding in free-text form~\citep{gao-etal-2023-precise} and in structured form, the latter a single factorized hypothesis through the same renderer as FHS; parallel sampling~\citep{wang-etal-2023-selfconsistency}, drawing $J{=}2$ i.i.d.\ samples, the same number of hypotheses FHS draws and, like FHS, from an unmodified generator rather than one prompted to differ; decomposed retrieval~\citep{fang-glass-2026-beyond}; and FHS itself. \textbf{Iterative and learned:} intrinsic self-refinement~\citep{zhang-etal-2025-understanding,yang-etal-2025-confidence}, retrieval-feedback refinement~\citep{yoon-etal-2025-ask,wang-etal-2025-aelc}, and FHS-Seq (Section~\ref{sec:negative}). All grounding methods share one backbone, index, tokenizer, and the final listwise selector. Every baseline is an \emph{adaptation}: the methods we compare against assume an explicit query, which this task does not provide. Appendix~\ref{app:baselines} states for each one what the cited method assumes, why our input does not satisfy it, and what our instantiation does instead.

We categorize compared methods into two paradigms:
\textbf{(1) Single-pass and parallel:} direct retrieval \citep{wang-etal-2025-fintagging}; free-text \citep{gao-etal-2023-precise} and structured single-pass grounding (the latter rendering a single factorized hypothesis); parallel sampling \citep{wang-etal-2023-selfconsistency} with $J{=}2$ i.i.d.\ samples; decomposed retrieval \citep{fang-glass-2026-beyond}; and \textbf{FHS}. 
\textbf{(2) Iterative and learned:} intrinsic self-refinement \citep{zhang-etal-2025-understanding,yang-etal-2025-confidence}, retrieval-feedback refinement \citep{yoon-etal-2025-ask,wang-etal-2025-aelc}, and \textbf{FHS-Seq} (Section~\ref{sec:negative}). All methods share an identical backbone model, index, tokeniser, and final listwise selector. Because prior methods assume an explicit input query (which our task setting lacks) each baseline represents an adaptation; exact mappings and adaptations are detailed in Appendix~\ref{app:baselines}

\subsection{Experimental Setting}
\label{sec:settings}

To control for model capacity, all LLM-based operations across all methods use
Qwen3-32B, including grounding, hypothesis generation, refinement,
candidate-level verification, and other baseline-specific operations. After
retrieval, every method additionally passes its candidates to the same
Qwen3-32B listwise selector. Full configurations are
provided in Appendix~\ref{app:implementation}. Recall and MRR are measured at the end of retrieval, before the shared listwise
selector. Acc.\ is top-1 accuracy after that selector and is the only metric
measured downstream of it. Each performance value is the mean of three runs on
the test set, and std is the standard deviation across those runs.
Appendices~\ref{app:method_specification} and~\ref{app:experimental_protocol}
give implementation details, baseline adaptation, configuration selection on a disjoint development sample, and the evaluation protocol. Appendix~\ref{app:retriever_robustness} repeats the study with dense and hybrid retrieval. Appendix~\ref{subsec: cost} reports computational cost. Appendix~\ref{app:cases} shows the case study.

\begin{table}[!t]
\centering
\setlength{\tabcolsep}{2.5pt}
\scriptsize
\resizebox{\columnwidth}{!}{%
\begin{tabular}{@{}p{2.05cm}cccccc@{}}
\toprule
& \multicolumn{4}{c}{\textbf{Retrieval}} & \multicolumn{2}{c}{\textbf{Final}} \\
\cmidrule(lr){2-5}\cmidrule(lr){6-7}
\textbf{Method} & \textbf{R@1} & \textbf{R@10} & \textbf{R@50} & \textbf{MRR} & \textbf{Acc.} & \textbf{std} \\
\midrule
  Gold oracle & 0.955 & 1.000 & 1.000 & 0.972 & --- & --- \\
\multicolumn{7}{@{}l}{\emph{Single-pass and parallel}} \\
  Direct retr.\newline{\scriptsize\citep{wang-etal-2025-fintagging}} & 0.041 & 0.240 & 0.562 & 0.111 &
  0.132 & 2.3e-4 \\
  One-pass, free-text\newline{\scriptsize\citep{gao-etal-2023-precise}} & 0.106 & 0.304 & 0.496 & 0.171
  & 0.199 & 2.3e-4 \\
  One-pass, structured & 0.122 & 0.372 & 0.554 & 0.203 & 0.226 & 6.2e-3 \\
  Parallel, stochastic\newline{\scriptsize$J{=}2$, \citep{wang-etal-2023-selfconsistency}} & 0.116 &
  0.370 & 0.583 & 0.194 & 0.228 & 1.8e-3 \\
  Decomposed\newline{\scriptsize\citep{fang-glass-2026-beyond}} & 0.102 & 0.307 & 0.478 & 0.166 & 0.213
  & $0$ \\
  \textbf{FHS (full)} & \textbf{0.185} & 0.397 & 0.543 & \textbf{0.257} & \textbf{0.255} & 1.5e-3
  \\
  \midrule
  \multicolumn{7}{@{}l}{\emph{Iterative and learned}} \\
  Intrinsic refine.\newline{\scriptsize\citep{zhang-etal-2025-understanding}} & 0.126 & 0.412 & 0.620 &
  0.216 & 0.234 & 4.3e-3 \\
  Feedback refine.\newline{\scriptsize\citep{yoon-etal-2025-ask,wang-etal-2025-aelc}} & 0.141 &
  \textbf{0.422} & \textbf{0.660} & 0.238 & 0.223 & 2.3e-4 \\
  FHS-Seq & 0.112 & 0.362 & 0.544 & 0.193 & 0.226 & --- \\
\bottomrule
\end{tabular}}
\caption{Main results on Financial Tagging. All methods share the same index, backbone, and downstream selector. Baseline details appear in Appendix~\ref{app:baselines}. Bold marks the best value in each column.}
\label{tab:main_results}
\label{tab:seq_results}
\end{table}

\subsection{Main Results}
FHS gives the strongest head-ranking performance on Financial Tagging
(Table~\ref{tab:main_results}). Its Recall@1 is 0.185, an absolute gain of
0.044 over retrieval-feedback refinement, the strongest baseline at rank one.
It also improves MRR from 0.238 to 0.257 and final accuracy from the
best baseline value of 0.234 to 0.255. The ordering changes deeper in the
ranking: retrieval-feedback refinement reaches 0.660 Recall@50, compared
with 0.543 for FHS. Thus, iterative accumulation expands the set of retrieved
candidates, but its additional depth does not produce a better ordering at the
head or a better final prediction. The same head-ranking advantage transfers to CodiEsp (Table~\ref{tab:codiesp_main_results}). Relative to the strongest baseline
for each metric, FHS improves Recall@1 from 0.201 to 0.264, MRR from 0.298
to 0.352, and final accuracy from 0.322 to 0.330. Unlike on Financial
Tagging, FHS also achieves the highest Recall@10 and Recall@50. Structured
one-pass grounding is the strongest baseline on this domain, whereas both
iterative methods perform worse. Across the two domains, the consistent gain
is therefore at the head of the ranking; deeper retrieval gains from iteration
are domain-dependent and do not reliably improve the final decision.
Additional retriever, modality, and cost analyses appear in
Appendix~\ref{app:test_dia}.
  \begin{table}[!t]
  \centering
  \setlength{\tabcolsep}{2.5pt}
  \scriptsize
  \resizebox{\columnwidth}{!}{%
  \begin{tabular}{@{}p{2.05cm}cccccc@{}}
  \toprule
  & \multicolumn{4}{c}{\textbf{Retrieval}} & \multicolumn{2}{c}{\textbf{Final}} \\
  \cmidrule(lr){2-5}\cmidrule(lr){6-7}
  \textbf{Method} & \textbf{R@1} & \textbf{R@10} & \textbf{R@50} & \textbf{MRR} & \textbf{Acc.} &
  \textbf{std} \\
  \midrule
  Gold oracle & 0.998 & 1.000 & 1.000 & 0.999 & 0.998 & -- \\
  \midrule
  \multicolumn{7}{@{}l}{\emph{Single-pass and parallel}} \\
  Direct retr. & 0.046 & 0.174 & 0.314 & 0.087 &
  0.203 & 1.8e-4 \\
  One-pass, free-text & 0.135 & 0.357 & 0.553 & 0.211
  & 0.242 & 1.8e-4 \\
  One-pass, structured & 0.201 & 0.487 & 0.626 & 0.298 & 0.322 & 6.7e-4 \\
  Parallel, stochastic & 0.165 &
  0.412 & 0.601 & 0.246 & 0.270 & 7.9e-4 \\
  Decomposed & 0.160 & 0.372 & 0.495 & 0.232 & 0.280
  & 1.8e-4 \\
  \textbf{FHS (full)} & \textbf{0.264} & \textbf{0.502} & \textbf{0.661} & \textbf{0.352} &
  \textbf{0.330} & 1.02e-3 \\
  \midrule
  \multicolumn{7}{@{}l}{\emph{Iterative and learned}} \\
  Intrinsic refine. & 0.161 & 0.413 & 0.595 &
  0.244 & 0.253 & 1.8e-4 \\
  Feedback refine. & 0.144 & 0.393 &
  0.583 & 0.227 & 0.238 & 1.8e-4 \\
  \bottomrule
  \end{tabular}}
  \caption{Main results on CodiEsp. All methods share the same index and downstream selector. Bold marks the best value in each column.}
  \label{tab:codiesp_main_results}
  \end{table}
\subsection{Ablation Studies}
% Table~\ref{tab:ablation} shows that the main components affect different
% parts of the ranking. 
% Replacing the factorized path with a same-size
% free-text ensemble reduces Recall@1 from 0.185 to 0.116 and MRR

% We use \emph{factorized path} for the factorized hypotheses together
% with the verifier that reads them. (Table~\ref{tab:ablation}) Replacing it with a same-size
% free-text ensemble reduces Recall@1 from 0.185 to 0.116 and MRR from
% 0.257 to 0.194, though Recall@50 increases from 0.543 to 0.583.

We use \emph{factorized path} for the factorized hypotheses together
with the verifier that reads them. Replacing it with a same-size
free-text ensemble reduces Recall@1 from 0.185 to 0.116 and MRR from
0.257 to 0.194, though Recall@50 increases from 0.543 to 0.583
(Table~\ref{tab:ablation}).
This row is run without the candidate-level verifier. Restoring the same verifier to the free-text ensemble recovers only part of the loss,
raising Recall@1 to 0.148 and MRR to 0.221. With the verifier held
fixed, factorization therefore adds 0.037 in Recall@1 and 0.036 in
MRR, against 0.008 and 0.011 without it. Conversely, the same verifier
gains 0.032 in Recall@1 on free text but 0.061 on factorized
hypotheses. Most of factorization's benefit thus comes from its interaction with
verification, not from factorization alone.

By comparison, using only one factorized
hypothesis produces much smaller reductions of 0.010 in Recall@1 and 0.011
in MRR. 
Multiple hypotheses therefore provide a complementary gain, while the
larger contrast is with the free-text ensemble.
% Multiple hypotheses therefore provide a complementary gain, while
% the larger contrast is between the structured and free-text paths. 
The rendering and verifier ablations further locate where the gains arise.
Removing the definition-style query causes larger losses than removing the
label form, reducing Recall@1 by 0.034 and final accuracy by 0.036. The
candidate-level verifier increases Recall@1 from 0.124 to 0.185 and MRR
from 0.205 to 0.257, while leaving Recall@50 unchanged at 0.543. It
therefore improves the ordering of an existing candidate pool rather than
expanding its coverage, consistent with FHS's focus on head ranking. In
contrast, replacing summed with mean RRF changes Recall@1 by only 0.003
and final accuracy by 0.004, indicating that the main gains do not depend
on the particular RRF aggregation convention. The label-coverage row
evaluates a shared index component rather than an FHS-specific mechanism:
its large drop shows that all methods benefit from aligning queries with the
taxonomy's compositional labels. Further design and sensitivity analyses are
provided in Appendices~\ref{app:test_dia} and \ref{app:design}.
\begin{table}[t]
\centering
\setlength{\tabcolsep}{2.5pt}
\scriptsize
\resizebox{\columnwidth}{!}{%
\begin{tabular}{@{}p{2.05cm}cccccc@{}}
\toprule
& \multicolumn{4}{c}{\textbf{Retrieval}} & \multicolumn{2}{c}{\textbf{Final}} \\
\cmidrule(lr){2-5}\cmidrule(lr){6-7}
\textbf{Variant} & \textbf{R@1} & \textbf{R@10} & \textbf{R@50} & \textbf{MRR} & \textbf{Acc.} & \textbf{std} \\
\midrule
\textbf{FHS (full)} & \textbf{0.185} & \textbf{0.397} & 0.543 & \textbf{0.257} & \textbf{0.255} & 1.5e-3 \\
\midrule
\multicolumn{7}{@{}l}{\emph{Generation}} \\
$-$ ensemble & 0.175 & 0.382 & 0.534 & 0.246 & 0.243 & 4.0e-4 \\
$-$ factorized path & 0.116 & 0.370 & \textbf{0.583} & 0.194 & 0.228 & 1.8e-3 \\
\quad $+$ verifier & 0.148 & 0.381 & 0.583 & 0.221 & 0.241 & 4.0e-4 \\
\multicolumn{7}{@{}l}{\emph{Rendering}} \\
$-$ label-form & 0.185 & 0.389 & 0.535 & 0.254 & 0.246 & 1.4e-3 \\
$-$ def-form & 0.151 & 0.348 & 0.492 & 0.218 & 0.219 & 4.0e-4 \\
\multicolumn{7}{@{}l}{\emph{Aggregation (substitutions)}} \\
mean RRF & 0.182 & 0.394 & 0.537 & 0.250 & 0.251 &  1.7e-3\\
raw fused scores & 0.176 & 0.388 & 0.535 & 0.249 & 0.246 & 1.5e-3 \\
\multicolumn{7}{@{}l}{\emph{Index (all methods)}} \\
$-$ label-cov & 0.101 & 0.266 & 0.423 & 0.163 & 0.196 & 1.5e-3 \\
% \midrule
\multicolumn{7}{@{}l}{\emph{Verifier}} \\
$-$ LLM verifier & 0.124 & 0.379 & 0.543 & 0.205 & 0.240 & 7.5e-4  \\
% Program-driven score & 0.097 & 0.385 & 0.550 & 0.188 & 0.240 & 0.016 \\
\midrule
Oracle choice & 0.222 & 0.437 & 0.592 & 0.295 & 0.262 & -- \\
\bottomrule
\end{tabular}}
\caption{Component ablations on Financial Tagging. ``$-$'' denotes
removal. The Aggregation rows are substitutions. The indented row adds the verifier back to the free-text ensemble. Bold marks the best non-oracle value. See
Appendix~\ref{app:readingablation} for row definitions.}
\label{tab:ablation}
\end{table}

\section{Conclusion}

We studied evidence-to-taxonomy retrieval, where observed evidence must be
interpreted before it becomes a useful retrieval query. FHS represents
competing interpretations as factorized hypotheses, fuses their retrieval
results, and reranks the resulting candidates. Across financial and clinical
coding, FHS achieves the strongest head-ranking and final prediction results
among the compared methods. Ablations attribute these gains primarily to the factorized hypothesis path (hypotheses + verifier) and definition-form rendering. These findings support
grounding through structured comparison among competing interpretations rather
than repeated refinement of a single query.

\section*{Limitations}

% FHS represents interpretations through a small set of semantic
% dimensions. In both evaluated domains, these dimensions are derived
% from existing taxonomy structure and metadata. Applying FHS to an
% inventory without such structure may require defining or inducing an
% appropriate schema. Automating this schema induction is a useful
% direction for future work.

FHS represents interpretations through a small set of semantic
dimensions. In both evaluated domains, these dimensions are derived
from existing taxonomy structure and metadata. The dimension names
are shared across domains, but their semantics are not. What transfers is the factorization mechanism, with
domain-specific slots that remain meaningful and verifiable against
the taxonomy. Applying FHS to an inventory without such structure may
require defining or inducing an appropriate schema. Automating this
schema induction is a useful direction for future work.

All LLM-based components in our experiments use Qwen3-32B. This
controlled setup isolates differences in representation and control
flow, but it does not establish how performance varies across model
families or scales. Future work should evaluate different model
families and scales to characterize the robustness--cost trade-off. 

Finally, the candidate-level verifier reorders the pooled candidates and cannot recover a gold concept that no hypothesis retrieved. We target this design at the diagnosis of Section 3.3, where the difficulty lies in head ranking rather than coverage. Settings where coverage is the bottleneck fall outside this diagnosis and are left to future work.

% FHS represents interpretations through a small set of semantic
% dimensions. In both evaluated domains, these dimensions are derived
% from existing taxonomy structure and metadata. Applying FHS to an
% inventory without such structure may require defining or inducing an
% appropriate schema. Automating this schema induction is a useful
% direction for future work.

% All LLM-based components in our experiments use Qwen3-32B. This
% controlled setup isolates differences in representation and control
% flow, but it does not establish how performance varies across model
% families or scales. Future work should evaluate larger
% backbones to characterize the robustness--cost trade-off.

\section*{Ethical Considerations}

% Both corpora are public and already de-identified at the source: the financial data comes from public company filings, and CodiEsp consists of clinical case reports drawn from open-access published literature (CC-BY 4.0), selected by a practicing physician and a clinical documentalist. We collect no new personal data, make no attempt to re-identify individuals, and release only code, prompts, vocabularies, and evaluation/diagnostic artifacts; we do not redistribute the corpora themselves and instead point to the original CC-BY 4.0 release.

The study uses public financial reports and the publicly released CodiEsp corpus, which was constructed from publicly available clinical case reports. We do not collect new personal
data or attempt to identify individuals. Automated taxonomy tagging may
nonetheless affect downstream financial analysis, clinical coding, and
reporting. FHS should therefore support candidate generation and expert
decision making rather than replace human review. Incorrect grounding may
propagate into subsequent analyses or decisions. We recommend retaining
auditable hypothesis trajectories and candidate-level support scores, and
periodically reviewing performance across concept families and evidence
types.

Neither corpus is user-generated: the financial data consists of public company filings and the CodiEsp cases are published, physician-selected case reports, so we did not find offensive content in the evidence side. The ICD-10-CM inventory, however, retains a number of legacy clinical terms in its inclusion notes and index entries that would be considered stigmatizing today. We index these strings verbatim because retrieval fidelity to the official code descriptions requires it, and we neither generate nor paraphrase such terms in system output.

\section*{Artifact Use and Licensing}
\label{sec:artifact-license}

Both corpora are public and already de-identified at the source: the financial data comes from public company filings, and CodiEsp consists of clinical case reports drawn from open-access published literature (CC-BY 4.0), selected by a practicing physician and a clinical documentalist. We collect no new personal data, make no attempt to re-identify individuals, and release only code, prompts, vocabularies, and evaluation/diagnostic artifacts; we do not redistribute the corpora themselves and instead point to the original CC-BY 4.0 release.

\section*{Use of AI Assistants}

The authors used AI assistants, including ChatGPT, to support language polishing, brainstorming, and LaTeX editing. All technical claims, experimental results, analysis, and final writing decisions were reviewed and verified by the authors.

% Our experiments use public-company filings retrieved from SEC EDGAR, released benchmark metadata from FinAuditing, and publicly available US-GAAP taxonomy releases. We use these resources for research evaluation and cite their original sources. We do not redistribute raw SEC filing packages or proprietary model outputs as part of this paper. Any released code or derived data will include references to the original resources and follow the applicable terms of use of the source artifacts.

% The fields used in our experiments contain company identifiers, filing metadata, audit queries, taxonomy concepts, reporting periods, and numerical ground-truth answers; they do not contain private personal information or user-generated offensive content. Public SEC filings may include publicly disclosed officer or signatory names, but these are not used as prediction targets or evaluation labels.

% codieps is Creative Commons Attribution 4.0 International. publicly available US-GAAP taxonomy releases. 

% \section*{Acknowledgments}
% Omitted for anonymous review.

\bibliography{custom}
\clearpage
\appendix

\section{Task Instantiation Details}
\label{app:instantiation}

\subsection{Financial Tagging}
\label{app:fintagging}

\begin{table}[!htbp]
\centering
\small
\begin{tabular}{@{}lrr@{}}
\toprule
\textbf{Statistic} & \textbf{Table} & \textbf{Text} \\
\midrule
Source contexts & 110 & 81 \\
Target facts & 2,341 & 168 \\
Unique concepts & 327 & 88 \\
Facts per context & 21.28 & 2.07 \\
\bottomrule
\end{tabular}
\caption{Test split statistics.}
\label{tab:data_stats}
\end{table}

\paragraph{Splits and statistics.}
The test split has 191 source contexts, 110 containing tables and 81 pure
narrative passages, yielding \num{2509} target facts over 388 distinct gold
concepts (Table~\ref{tab:data_stats}; the per-modality concept counts sum to
more than 388 because 27 concepts occur in both modalities). Datatypes are
$90.0\%$ monetary, $4.5\%$ percent, $2.8\%$ shares, $2.3\%$ per-share and
$0.4\%$ integer. Configuration selection used a disjoint 661-fact development
sample of 70 contexts drawn from the training split
(Appendix~\ref{app:devsample}); contexts, not facts, were sampled, and the two
sets share no source context.

\paragraph{Locus and context serialization.}
The locus $\ell_i$ carries the target cell's row header, column header, and
declared datatype for tabular facts, and the mention span for narrative ones.
Row and column headers are pipe-delimited header chains, and the source context
$X_i$ is the benchmark's own HTML \texttt{<table>} markup, passed through with
whitespace normalized. Serialization is capped at \num{12000} characters, and
the rendered prompt at \num{16000} tokens; when a prompt exceeds the token
budget the character cap backs off through \num{8000}, \num{5000} and \num{2500}
until it fits. Truncation preserves both ends and elides the middle, so the
opening headers and the closing totals rows both survive. Narrative contexts are
passed whole under the same budget. Every method receives the same
serialization, and direct retrieval uses it verbatim as its query, so this
choice sets the floor as well as the input to grounding.

\paragraph{Taxonomy filtering and index.}
We index all \num{17388} concepts of the 2024 US-GAAP taxonomy without
filtering: no concept is removed for being abstract, deprecated, or lacking a
documentation string, and \num{2470} of the \num{17388} carry no documentation
text. We index the \textsf{us-gaap} namespace only; every gold tag in the test
split is a \textsf{us-gaap} concept, so no company-specific extension element
appears as a target. Each concept is one document holding its canonical label,
its official documentation string where one exists, and its declared datatype.
Retrieval is BM25 at depth $K{=}200$, implemented in-repo rather than through a
library, with $k_1{=}1.5$, $b{=}0.75$ and the Lucene-style
$\log(1+(N-\mathrm{df}+0.5)/(\mathrm{df}+0.5))$ idf.

\paragraph{Datatype pre-filter.}
Before scoring, each query is restricted to the concepts whose datatype is
compatible with the fact's own, so a monetary cell is never scored against a
share-count concept. Compatibility is exact match on the declared datatype, with
a fall back to the full index for a datatype the taxonomy does not declare; no
test fact triggers that fall back. The filter leaves a mean of \num{6919}
candidates per query against \num{17388} unfiltered, and retains the gold
concept on $100.0\%$ of test facts, so it removes roughly $60\%$ of the pool at
no cost in reach. The filter is a property of the index and applies identically
to every method compared, direct retrieval included.

\paragraph{Label-coverage term.}
Sparse scoring over compositional labels is length-biased: a short generic
concept loses to a longer label that merely contains the query's terms. The
retriever therefore adds to the range-normalized BM25 score two token-coverage
terms between the query and the candidate's canonical label, one normalized by
the label's length and one by the query's, under a single weight
$w_{\mathrm{cov}}{=}1.0$. Appendix~\ref{app:coverage} gives both terms, the
shared tokenizer, and the length breakdown of the gain. Like the datatype
filter, the term is a property of the index rather than an FHS component and is
enabled identically for every method; Table~\ref{tab:ablation} reports what
removing it costs.

\paragraph{Vocabularies and the match operator.}
The four controlled vocabularies (18 \textsc{Qualifier}, 7 \textsc{Scope},
11 \textsc{Temporal}, and 14 \textsc{Family} categories) were derived from
the taxonomy's structural metadata and label conventions and fixed on the
development sample before test evaluation. A normalization map projects the
generator's free-form outputs onto these vocabularies. Values that cannot be
normalized are logged for diagnostic analysis but do not modify the
vocabularies during evaluation. For the two vocabulary-free dimensions,
\textsc{Role} and \textsc{Event}, a value matches a candidate when at least
half of its normalized content tokens occur in the candidate's concatenated
label and documentation text. 

The four vocabulary-matched dimensions define a candidate's
\emph{category profile}, represented by the tuple of its normalized
\textsc{Family}, \textsc{Qualifier}, \textsc{Scope}, and \textsc{Temporal}
values. Candidates with different profiles represent distinct structural
interpretations along at least one of these dimensions. The verifier window
in Section~\ref{sec:llmverifier} therefore retains the highest-ranked
candidate from each distinct profile, increasing structural diversity within
the window and reducing redundancy among candidates that share the same
profile.

\paragraph{Evaluation convention.}
A prediction is correct when it matches the gold concept's identifier exactly
after stripping the \textsf{us-gaap:} namespace prefix; identifiers are
canonical camel-case, so no case folding is applied. Recall and MRR are computed
over the retrieved list at depth $K{=}200$ before the shared listwise selector;
accuracy is top-1 of the selector's output.

\paragraph{Shared backbone and selector.}
All grounding methods use \textsc{Qwen3-32B} as the generator and, where
applicable, the verifier backbone, with structured decoding. Decoding is greedy
for every method except the two that need sample diversity: FHS draws its $J$
hypotheses at temperature $0.8$, as does parallel sampling. Downstream of every
method a shared listwise selector, the same \textsc{Qwen3-32B} model, receives
that method's top-$K$ list and returns at most twenty ranked concepts. It is
not part of any method's contribution; Appendix~\ref{app:protocol} states what
else is held identical across arms.

\subsection{CodiEsp Diagnosis Coding}
\label{app:codiesp}

\paragraph{Splits and statistics.}
The evaluation set is a deterministic exact-relocation slice of the official
CodiEsp test split, not a new random split. We keep all 250 test clinical
cases. From the diagnosis annotations, after inventory filtering,
deduplication, and exact English mention relocation, the split contains
\num{3144} target facts over \num{958} distinct gold codes
(Table~\ref{tab:codiesp_data_stats}). The full prepared diagnosis set contained
\num{3431} facts; \num{287} were dropped because they did not pass the exact
English relocation criterion. No development split or CodiEsp-specific
configuration selection is used.

\begin{table}[!htbp]
\centering
\small
\begin{tabular}{@{}lr@{}}
\toprule
\textbf{Statistic} & \textbf{CodiEsp-D} \\
\midrule
Source documents & 250 \\
Target facts & 3,144 \\
Unique gold codes & 958 \\
Facts per document & 12.58 \\
3-character gold codes & 410 \\
4--5-character gold codes & 2,547 \\
6--7-character gold codes & 187 \\
\bottomrule
\end{tabular}
\caption{CodiEsp diagnosis test statistics after exact English mention
relocation.}
\label{tab:codiesp_data_stats}
\end{table}

\paragraph{Locus and context serialization.}
The locus $\ell_i$ carries the relocated English diagnosis mention and the
local English clinical locus. Because CodiEsp offsets are offsets into Spanish
clinical text, each gold mention is first re-located in the English
machine-translated context with one model call and an exact-substring check.
The retained split has \texttt{relocation\_parse\_ok\_rate}=1.0 and
\texttt{relocation\_exact\_substring\_rate}=1.0; among retained facts,
\num{904} relocations were found from an aligned-sentence candidate scope and
\num{2240} from a document-level candidate scope before exact substring
selection. The source context $X_i$ is serialized as clinical mention, code
class \texttt{diagnosis}, input type, and English source context, with the
runner's default \texttt{12000}-character context budget for query generation
and reranking. Every method receives the same serialization, and direct
retrieval uses it verbatim as its query, so this choice sets the floor as well
as the input to grounding.

\paragraph{Taxonomy filtering and index.}
Of the local diagnosis-code labels available to the CodiEsp setup, we retain
\num{71344} candidates after intersecting the CodiEsp valid diagnosis-code list
with billable FY2018 diagnosis codes. Each code is one document holding its
identifier, canonical English label, documentation text, and structural
metadata. Since the inventory has no single official definition paragraph per
code, the documentation string concatenates, in order, current-code inclusion
terms, inherited include notes, hierarchy path text, and alphabetic-index
lead/sub-term paths resolving to the code. Exclusion notes are kept separate
and are not included in retrieval text. Retrieval is BM25 at depth $K{=}200$,
implemented by the local runner with $k_1=1.5$ and $b=0.75$.

\paragraph{Datatype pre-filter.}
The analogue of the financial datatype pre-filter is the diagnosis-code
inventory restriction. Every fact in this CodiEsp instantiation has code class
\texttt{diagnosis}, and every retained candidate is a billable diagnosis code,
so the shared type-filter hook restricts scoring to that diagnosis inventory.
All retained facts have gold codes in the indexed diagnosis inventory; no
  method-specific denominator filtering is applied during evaluation. The
filter is a property of the index and applies identically to every method
compared, direct retrieval included.

\paragraph{Label-coverage term.}
Sparse scoring over compositional diagnosis labels can favor longer labels that
repeat query terms without matching the intended code reading. The retriever
therefore supports adding to the range-normalized BM25 score two token-coverage
terms between the query and the candidate's canonical label, one normalized by
the label's length and one by the query's, under a single weight
$w_{\mathrm{cov}}$. Table~\ref{tab:codiesp_main_results} reports the
$w_{\mathrm{cov}}=1.0$ cell.

\paragraph{Vocabularies and the match operator.}
The five controlled vocabularies were fixed before test evaluation:
21 \textsc{Family} categories, 9 \textsc{Role} classes, 21
\textsc{Qualifier} modifiers, 5 \textsc{Scope} laterality values, and
12 \textsc{Temporal} encounter or extension values. \textsc{Family}
represents broad chapter-level clinical families; \textsc{Role}
distinguishes classes such as disease/disorder, neoplasm,
injury/poisoning, external cause, and health-status factor;
\textsc{Scope} represents right, left, bilateral, unspecified-side, or
not-applicable status; and \textsc{Temporal} captures distinctions such
as encounter type, sequela, healing status, stage, and fetus-specific
extensions. A normalization map projects the generator's free-form
outputs onto these controlled values. Values that cannot be normalized
are logged for diagnostic analysis but do not modify the vocabularies
during evaluation. For the vocabulary-free \textsc{Event} dimension, a
value matches a candidate when its normalized content tokens overlap
the candidate's concatenated label and documentation text.

The five vocabulary-matched dimensions define a candidate's
\emph{category profile}, represented by the tuple of its normalized
\textsc{Family}, \textsc{Role}, \textsc{Qualifier}, \textsc{Scope}, and
\textsc{Temporal} values. Codes sharing a profile differ primarily in
the specific condition expressed by \textsc{Event}, whereas different
profiles capture broader distinctions in clinical family, code class,
modifier, laterality, or encounter status. As in Financial Tagging, the
verifier window retains the highest-ranked candidate from each distinct
profile, increasing structural diversity within the window and reducing
redundancy among candidates with the same profile.

\paragraph{Evaluation convention.}
A prediction is correct when it matches the gold diagnosis-code identifier
exactly after the runner's tag normalization. Each fact has exactly one gold
code. Recall@10, Recall@50, Recall@200, and MRR are computed over the retrieved
list at depth $K{=}200$ before the shared listwise selector. Accuracy is the
top-1 output of the selector.

\paragraph{Shared backbone and selector.}
All grounding methods use \textsc{Qwen3-32B} as the generator and, where
applicable, the verifier backbone. Downstream of every method with reranking
enabled, a shared listwise selector using the same \textsc{Qwen3-32B} model
receives that method's top-$K$ list and returns at most twenty ranked codes. It
is not part of any method's contribution; the same selector prompt, candidate
format, and evaluation code are held fixed across arms.

\section{Method Specification}
\label{app:method_specification}
\subsection{Pseudocode}
\label{app:floats}

Algorithm~\ref{alg:ags} states the deployed method in full: the $J$ hypotheses, the two renderings per hypothesis, the consensus fusion, and the candidate-level verifier that reorders the head before the shared listwise selector.

\begin{algorithm}[t]
\caption{FHS: parallel factorized-hypothesis grounding}
\label{alg:ags}
\begin{algorithmic}[1]
\small
\Require located fact $x$, taxonomy $\mathcal T$, generator $G$,
label-form renderer $\rho$, retriever $R$, verifier $V$, samples $J$,
retrieval depth $K$, candidate window $K_v$

\State $\mathcal R\gets\varnothing$ \Comment{ranking set}

\For{$j=1$ \textbf{to} $J$} \Comment{parallelizable}
    \State $(h_j,q_j^{\mathrm{def}})\gets G(x)$
    \Comment{dimensions, $\bot$ allowed, and definition-form query}
    \State $q_j^{\mathrm{lab}}\gets\rho(h_j)$
    \Comment{rendered from resolved dimensions}
    \State $\mathcal R_j\gets
    \{R(q_j^{\mathrm{def}},\mathcal T,K)\}$

    \If{$x$ is tabular \textbf{and}
    $q_j^{\mathrm{lab}}\neq\varnothing$}
        \State $\mathcal R_j\gets\mathcal R_j
        \cup\{R(q_j^{\mathrm{lab}},\mathcal T,K)\}$
    \EndIf

    \State $\mathcal R\gets\mathcal R\cup\mathcal R_j$
\EndFor

\State $\mathcal U\gets
\bigcup_{\mathcal C\in\mathcal R}\mathcal C$
\State $S\gets\textsc{SumRRF}(\mathcal R)$
\Comment{Eq.~\ref{eq:sumrrf}}
\State $\widetilde S\gets\textsc{RangeNorm}(S,\mathcal U)$
\Comment{Eq.~\ref{eq:normalize}}

\For{$j=1$ \textbf{to} $J$} \Comment{parallelizable}
    \State $\mathcal B_j\gets
    \textsc{ProfileRepresentatives}(\mathcal U,\widetilde S,K_v)$
    \Comment{profile-diverse window with rank fallback}
    \State $\{v_j(c):c\in\mathcal B_j\}\gets
    V(x,h_j,\mathcal B_j)$
    \Comment{dimension-level verdicts for each candidate}
\EndFor

\For{$c\in\mathcal U$}
    \State $\overline v(c)\gets
    \textsc{AggregateSupport}
    (c,\{\mathcal B_j,v_j\}_{j=1}^{J})$
    \Comment{Eq.~\ref{eq:llm_support}; window-mean fallback}
    \State $S_{\mathrm{final}}(c)\gets
    \widetilde S(c)+\beta\,\overline v(c)$
    \Comment{Eq.~\ref{eq:hybrid_score}}
\EndFor

\State \Return
$\operatorname{TopK}_{c\in\mathcal U}S_{\mathrm{final}}(c)$
\end{algorithmic}
\end{algorithm}

\subsection{The label-form renderer}
\label{app:render}

% The tokenizer $\mathrm{tok}(\cdot)$ shared by the renderer and the coverage term of Eq.~\ref{eq:coverage} drops function words, applies a light plural rule, and emits both the original string and its camel-case split, because taxonomy labels are written both ways (\texttt{AssetsHeldForSale} and \emph{assets held for sale}) and a query should match either. Duplicates are kept rather than collapsed, and unresolved dimensions are skipped rather than placeheld. The definition form is in practice compact rather than prose -- typically the resolved values joined by delimiters -- so the two renderings differ less in register than their names suggest; if the generator returns that field empty, the resolved values are joined as a fallback.

The tokenizer $\mathrm{tok}(\cdot)$ shared by the renderer and the
coverage term of Eq.~\ref{eq:coverage} drops function words, applies
a light plural rule, and emits both the original string and its
camel-case split, because taxonomy labels are written both ways
(\texttt{AssetsHeldForSale} and \emph{assets held for sale}) and a
query should match either. The renderer preserves repeated tokens,
but the query is deduplicated before retrieval scoring and both
arguments of Eq.~\ref{eq:coverage} are sets, so repeated values do
not receive additional weight. Unresolved dimensions are skipped
rather than placeheld.

\subsection{Generator output schema}

Each generator call returns the six dimensions, the literal string \texttt{UNRESOLVED} for any dimension the evidence does not support, and a \texttt{retrieval\_query} field holding the definition-form query of Section~\ref{sec:render}; that field is why the definition form costs no call of its own. The prompt also lists an operator vocabulary and the schema carries an \texttt{operators} field, inherited from the refinement arms that share this prompt. The renderer and every scoring path read only the dimensions and \texttt{retrieval\_query}, so that field is emitted and then unused by FHS; we keep it so the two arms issue the same prompt. Decoding is structured, and the $J$ hypotheses are drawn by stochastic sampling of one prompt. The prompt defines each dimension in one line and requires the sentinel \texttt{UNRESOLVED} rather than a guess where the evidence does not support a value. Generators do not see the taxonomy or any candidate concepts, so a hypothesis cannot be contaminated by the candidate list it will later be scored against.

\subsection{Fusion and Normalization}
\label{app:fusion}

Both operations are standard; we give them here for completeness. Let $r$ index the rankings produced across all hypotheses and renderings, with pool $\mathcal U=\bigcup_r\mathcal C_r$. Summed reciprocal rank fusion~\citep{cormack2009reciprocal} is
\begin{equation}
S(c)=\sum_{r\,:\,c\in\mathcal C_r}\frac{1}{\kappa+\operatorname{rank}_r(c)},
\label{eq:sumrrf}
\end{equation}
with $\kappa=60$. The sum runs only over the rankings that returned $c$, so a candidate found by several hypotheses accrues a term from each. Scores are then range-normalized over the pool,
\begin{equation}
\widetilde S(c)=\frac{S(c)-\min_{c'\in\mathcal U}S(c')}{\max_{c'\in\mathcal U}S(c')-\min_{c'\in\mathcal U}S(c')}.
\label{eq:normalize}
\end{equation}
Normalization matters for the weight $\beta$ of Eq.~\ref{eq:hybrid_score}: without it the same $\beta$ behaves inconsistently across ensemble sizes, since the raw fused range grows with the number of rankings. Table~\ref{tab:ablation} reports both substitutions, mean RRF for the sum and raw scores for the normalization.

\subsection{Implementation and Hyperparameters}
\label{app:implementation}

The experiments were run with PyTorch 2.7.1+cu126 on a single NVIDIA B200 GPU under Red Hat Enterprise
  Linux 9.6 (Plow). Table~\ref{tab:hyperparameters} lists every setting the deployed configuration pins; any value not shown there is a library default we did not change.
\begin{table}[!htbp]
\centering
\small
\begin{tabular}{@{}ll@{}}
\toprule
\textbf{Setting} & \textbf{Value} \\
\midrule
Retrieval depth $K$ & 200 \\
Hypotheses $J$ & 2 \\
Fusion variant & sum (Eq.~\ref{eq:sumrrf}) \\
Score scaling & range-normalized \\
Rerank weight $\beta$ & 0.6 \\
Candidate window $K_v$ & 10 \\
Window scan depth & 60 \\
Verified dimensions & all $M{=}6$ \\
Generator temperature & 0.8 \\
Coverage rescoring pool & full datatype pool \\
RRF constant $\kappa$ & 60 \\
Coverage weight $w_{\mathrm{cov}}$ & 1.0 \\
Rendering (table / text) & dual / definition \\
Candidate-level verifier & enabled \\
\midrule
\multicolumn{2}{@{}l}{\emph{Sequential control only}} \\
Rounds $B$ & 4 \\
\bottomrule
\end{tabular}
\caption{Hyperparameters. $J$ and $\beta$ were selected on the development sample of Appendix~\ref{app:devsample}; $K_v$ was fixed a priori and its sensitivity is reported in Table~\ref{tab:llm_window_sensitivity}. }
\label{tab:hyperparameters}
\end{table}

 All methods share the same retriever index, taxonomy serialization, tokenizer, and renderer. The candidate-level verifier, renderer, fusion, and rerank are each implemented once and invoked by every method and ablation through configuration flags, so no two reported numbers come from divergent code paths. We adopted this after two independently implemented aggregation routines produced inconsistent stage-decomposition values during development.

%% =====================================================================
\section{Experimental Protocol}
\label{app:experimental_protocol}
\subsection{Baseline Adaptations and Experimental Protocol}
\label{app:baselines}

Every baseline here is an \emph{adaptation}. The methods we compare against were designed for a
setting where an explicit query exists; our input is a located fact inside a shared context, so none
of them can be run unmodified. This appendix states, for each one, what the cited method assumes,
why our input does not satisfy that assumption, and what our instantiation does instead. We
describe the prompts as implemented, not as idealised.

\paragraph{Direct retrieval}~\citep{wang-etal-2025-fintagging}. The benchmark's own baseline. The
located fact together with its row and column context is serialized and used as the query
verbatim. No model call, so it is the only arm with no prompt.

\paragraph{One-pass grounding, free-text}~\citep{gao-etal-2023-precise,wang-etal-2023-query2doc}.
HyDE and Query2Doc expand an \emph{existing} query into a pseudo-document, which is then embedded or
matched. There is no query here to expand. Our instantiation gives the model the serialized evidence
and asks for one retrieval-ready description, returned as a single structured field; that
description is the query. This keeps the mechanism the cited work proposes -- put a generated
document-like text on the query side -- while supplying the input our task actually has.

\paragraph{One-pass grounding, structured}. Ours, not adapted: a single factorized hypothesis
through the same renderer FHS uses, which separates the effect of the representation from the effect
of the ensemble.

\paragraph{Parallel sampling}~\citep{wang-etal-2023-selfconsistency}. Self-consistency samples one
prompt $N$ times and takes a majority vote over the answers. There is no answer to vote over: the
output is a query, so we fuse the $N$ \emph{retrievals} through the same aggregator FHS uses. We
report the arm that matches self-consistency's own procedure -- $N$ independent samples of one
unmodified prompt at the same decoding temperature FHS uses. A second variant, which tells the model
that this is sample $i$ of $N$ and to make each sample ``meaningfully distinct'', is a
forced-variation arm rather than i.i.d.\ sampling, and we report it separately for that reason.

\paragraph{Decomposed retrieval}~\citep{fang-glass-2026-beyond}. The cited line decomposes a complex
query into sub-queries retrieved separately. Our instantiation asks for one sub-query per semantic
dimension, retrieves each independently, and fuses them with the same aggregator. This is the
closest baseline to FHS that never forms a \emph{joint} reading of the fact: it splits the evidence
across dimensions instead of committing to a hypothesis over them, which is precisely the
distinction Section~\ref{sec:factorized} draws.

\paragraph{Intrinsic self-refinement}~\citep{zhang-etal-2025-understanding,yang-etal-2025-confidence}.
The cited work studies a model revising its own answer with no external signal. Ours keeps that
constraint exactly: each round the model is shown its previous interpretation and \emph{no retrieved
candidates}, and returns a critique together with a revised description.

\paragraph{Retrieval-feedback refinement}~\citep{yoon-etal-2025-ask,wang-etal-2025-aelc,asai-etal-2024-selfrag,shao-etal-2023-iterretgen}.
The Self-RAG / ITER-RETGEN loop alternates retrieval with generation over an explicit query. Ours
shows the model its previous interpretation together with the concepts retrieved by it, and asks
whether that interpretation is on the right track before rewriting it. The rewrite is free-form: the
prompt explicitly permits changing entity type, temporal scope, qualifiers, or aggregation level.

\paragraph{FHS-Seq}. Ours: identical to FHS in every component, including the candidate-level
verifier, and different only in control flow (Section~\ref{sec:negative}).

\subsection{What is held identical, what is matched, and what is not}
\label{app:protocol}

% \paragraph{Identical across every arm.} One backbone, one decoding configuration, one taxonomy
% index with the same coverage term, one tokenizer, one renderer where a renderer applies, one
% retrieval depth $K{=}200$, one fusion constant $\kappa{=}60$, one output-token budget, and the same downstream listwise selector.  The selector in particular is not part of any method's contribution: it runs after all of them, receives the pool from each, and returns at most twenty ranked tags which are placed above the rest of that method's ranking.

\paragraph{Identical across every arm.} All methods use the same backbone, output-token budget, taxonomy index, tokenizer, retrieval depth, and downstream selector. Decoding
is greedy except for methods that require sample diversity: FHS and
parallel sampling use stochastic decoding at temperature $0.8$. The selector in particular is not part of any method's contribution: it runs after all of them, receives the pool from each, and returns at most twenty ranked tags which are placed above the rest of that method's ranking.

% The selector runs after every method, receives that method's own
% top-$K$ candidates under the same input format and candidate budget,
% and returns at most twenty ranked tags, which are placed above the
% remainder of that method's ranking.

\paragraph{Matched on hypothesis count, not on model calls.} Every parallel arm draws $J{=}2$
samples, the number of hypotheses FHS draws. FHS then makes two further calls to verify, so it uses
four model calls against a parallel baseline's two. We report that gap rather than equalise it,
because equalising it would mean treating a verifier call as one generation call, and a verifier
call carries $K_v$ candidate labels and definitions -- it is much the larger of the two. Matching on
call count would therefore understate FHS's cost, not the baseline's.

\paragraph{Selected on development data.} $J$ and $\beta$ were chosen on the 661-fact development
sample of Appendix~\ref{app:devsample}; $K_v$ was fixed a priori and its sensitivity is reported
rather than tuned. The verified dimension set is not a tuned quantity either: it is the whole schema
the generator emits, so there is nothing to select over.

\paragraph{Measurement stage.} Recall and MRR are measured at the end of retrieval, before the
shared selector. Accuracy is top-1 after it, and is the only quantity measured downstream of it, so
a method that helps only by reordering what the selector already re-sorts shows a gain in the former
and little in the latter.

\subsection{Development Sample}
\label{app:devsample}

Configuration selection ($J$, $\beta$, fusion variant, rendering) used a frozen sample of 661 facts over 70 source contexts drawn from the training split, disjoint from test. Contexts were sampled rather than facts, preserving the grouping structure (Table~\ref{tab:devsample}; Table~\ref{tab:data_stats} gives the test split for comparison). 

\begin{table}[h]
\centering
\small
\begin{tabular}{@{}lrr@{}}
\toprule
\textbf{Statistic} & \textbf{Table} & \textbf{Text} \\
\midrule
Source contexts & 30 & 40 \\
Facts & 566 & 95 \\
Facts per context (mean) & 18.87 & 2.38 \\
Unique concepts & 131 & 55 \\
\bottomrule
\end{tabular}
\caption{[DEV] Development sample composition. }
\label{tab:devsample}
\end{table}
%Datatype mix: 88.7\% monetary, 6.5\% percent, 2.9\% shares, 2.0\% per-share.

Appendix tables marked [DEV] are computed on this sample under the unified index and are configuration-selection evidence only; every table reporting a final result is computed on the frozen test split. Confidence intervals are bootstrap, resampled at the source-context level, 2{,}000 iterations, with all contrasts paired per fact. Context-level resampling at $n=30$ tabular contexts yields wide intervals; this is a property of the development sample, not of the test protocol.

%% =====================================================================

\subsection{Reading the Component Ablation}
\label{app:readingablation}

Table~\ref{tab:ablation} mixes three kinds of row, and the distinction matters for what a delta
against FHS means.

\paragraph{Matched arms.} Six rows are the deployed method with exactly one field changed: FHS,
$-$ ensemble, $-$ label-form, $-$ definition-form, \emph{mean RRF} and \emph{raw fused scores}.
These rows use the same evaluation and the same shared
implementation, verifier, and judged dimensions. Where applicable,
they reuse the same generated hypotheses. Their candidate pools may
differ when the ablated component changes the issued queries or the
number of retrieved rankings. Such pool changes are part of the
component's causal effect. The one further
deviation is the rerank weight: $-$ label-form uses $\beta{=}0.8$ and $-$ definition-form
$\beta{=}0.2$ against $0.6$ elsewhere, because dropping a rendering halves the number of fused
rankings and $\beta$ is re-swept on the development sample whenever that count changes
(Appendix~\ref{app:beta}).

% All six score the same frozen candidate pool under the same $\mathrm{llm\_drop}$ verifier over the
% same six judged dimensions, so their gaps to FHS isolate the named component. 

% The one further
% deviation is the rerank weight: $-$ label-form uses $\beta{=}0.8$ and $-$ definition-form
% $\beta{=}0.2$ against $0.6$ elsewhere, because dropping a rendering halves the number of fused
% rankings and $\beta$ is re-swept on the development sample whenever that count changes
% (Appendix~\ref{app:beta}).

% \paragraph{Rows that cannot be matched.} Two rows differ from FHS in more than the named field,
% by construction rather than by oversight. \emph{$-$ verifier} sets $\beta{=}0$, so no verdict
% enters the score at all. \emph{$-$ factorization} is a free-text ensemble, which has
% no dimensions to verify, so it exchanges representation and verification together and bounds
% factorization's contribution rather than isolating it. 
\paragraph{Rows that cannot be matched.} Two rows differ from FHS in more than the named field,
by construction rather than by oversight. \emph{$-$ LLM verifier} sets $\beta{=}0$, so no verdict
enters the score at all. \emph{$-$ factorized path} is a free-text ensemble run with
$\beta{=}0$ as well, so it changes representation and verification together. The indented
\emph{$+$ verifier} row restores the deployed verifier to that ensemble, with the same prompt,
window, and $\beta{=}0.6$; the two rows together bound factorization's contribution with and
without verification.
\emph{$-$ label coverage} is a matched arm:
it removes the coverage term from FHS's own hypotheses and carries the same verifier over the same
six dimensions, so all of its reported columns come from one run.

\paragraph{The oracle row.} \emph{Oracle choice} carries the same verifier and the same six
judged dimensions as the matched arms, so its gap to FHS is a selection gap and not an artefact of the scoring path. Selecting the best hypothesis per fact produces a different retrieved
pool, reaching Recall@200 $0.734$, compared with $0.705$ for the
deployed FHS configuration. It bounds what a better selection rule
over this same hypothesis set could reach.

% It still differs in one respect the table shows: selecting the best hypothesis
% per fact means fusing each hypothesis separately, which reaches a slightly different pool
% (Recall@200 $0.734$ against $0.705$ for every other row). It bounds what a better selection rule
% over this same hypothesis set could reach.

\paragraph{Substitutions versus removals.} The two \emph{Aggregation} rows are named for what
replaces the deployed choice: \emph{mean RRF} averages the reciprocal-rank contributions instead
of summing them, and \emph{raw fused scores} feeds the verifier unnormalized scores instead of
range-normalized ones. Neither component can be deleted outright, several rankings must be
combined somehow, and the no-fusion case is exactly the $-$ ensemble ($J{=}1$) row.

%% =====================================================================
%appendix C
\section{Design Decisions on Development Data}
\label{app:design}

\subsection{Development Design Study}
\label{app:pilot}
\label{app:configabl}

\begin{table}[!htbp]
\centering
\small
\begin{tabular}{@{}lcccc@{}}
\toprule
\textbf{Arm} & \textbf{Cov@4} & \textbf{RRF} & \textbf{Rd-1} & \textbf{Jac.} \\
\midrule
Single hypothesis & 0.629 & --- & 0.629 & --- \\
Stochastic samples & \textbf{0.735} & 0.661 & 0.633 & 0.641 \\
Diversity-prompted & 0.664 & 0.555 & 0.555 & 0.499 \\
Dimension-directed & 0.682 & 0.610 & 0.604 & 0.536 \\
\bottomrule
\end{tabular}
\caption{[DEV] Hypothesis-generation pilot on tabular development data ($K=200$, four hypotheses). Plain stochastic sampling beats the forced-diversity variants.
\textbf{Cov@4} is the fraction of facts whose gold concept appears in the union of the four top-200 retrieved sets.
\textbf{RRF} is Recall@200 after summed reciprocal-rank fusion.
\textbf{Rd-1} is the mean Recall@200 over the four individual hypotheses.
\textbf{Jac.} is the mean pairwise Jaccard similarity among their top-200 retrieved sets.}
\label{tab:pilot}
\label{tab:configabl}
\end{table}

This section reports the development pilot used to choose the hypothesis-generation strategy before any test-split number was computed (Table~\ref{tab:pilot}). The pilot selects the sampling strategy; final component claims are evaluated separately on the test split in Table~\ref{tab:ablation}.

\paragraph{Plausibility before diversity.}
Plain stochastic sampling attains the highest accumulated coverage. Both variation-forcing arms reduce neighborhood overlap but also reduce coverage, because they lower single-hypothesis recall. These are development-pilot findings, not a universal claim that directed specialization cannot work.

\paragraph{Aggregation loss.}
For the selected stochastic-sampling configuration, accumulated coverage exceeds fused Recall@200 by 7.4 points. Reaching the correct neighborhood and consolidating it into a short list are therefore empirically separable.

We did not evaluate functionally specialized generators in place of stochastic samples (Section~\ref{sec:generators}). A matched-budget comparison should report each specialist's solo performance and selection frequency together with the ensemble result; the diversity pilot is why we leave the outcome open.

\subsection{Rerank Weight and Ensemble Size}
\label{app:beta}

The weight swept here is $\beta$, the weight on the reranking term added to the fused retrieval score $\widetilde S(c)$; Table~\ref{tab:beta} reports the sweep.

Range normalization is necessary to keep $\beta$ comparable across ensemble
sizes. Because summed RRF scales with the number of fused rankings whereas
$\overline{v}(c)$ remains in $[0,1]$, a fixed $\beta$ would otherwise assign
different relative weight to verifier support as $J$ changes.

\paragraph{Choice of $J$.}
At the deployed weight $\beta{=}0.6$, $J{=}2$ raises Recall@10 from $0.527$ to $0.541$ and MRR from $0.279$ to $0.295$ over $J{=}1$. $J{=}3$ adds $0.004$ MRR but lowers Recall@10 to $0.528$, at the cost of one more generation call and one more verification call. We therefore use $J{=}2$.

% Score normalization is necessary before tuning $\beta$. Without it the same weight behaves inconsistently across ensemble sizes. We therefore omit the redundant raw-score table and retain the normalized sweep used for selection.

\begin{table*}[!htbp]
\centering
\small
\begin{tabular}{@{}lccccccccc@{}}
\toprule
& \multicolumn{9}{c}{\textbf{$\beta$ (range-normalized)}} \\
\cmidrule(l){2-10}
\textbf{Metric / $J$} & \textbf{0} & \textbf{0.5} & \textbf{0.6} & \textbf{0.8} & \textbf{1.0} & \textbf{1.5} & \textbf{2.0} & \textbf{3.0} & \textbf{4.0} \\
\midrule
R@10, $J{=}1$ & 0.504 & 0.521 & \textbf{0.527} & 0.516 & 0.516 & 0.505 & 0.486 & 0.443 & 0.405 \\
R@10, $J{=}2$ & 0.502 & 0.537 & \textbf{0.541} & \textbf{0.541} & 0.537 & 0.518 & 0.498 & 0.461 & 0.443 \\
R@10, $J{=}3$ & 0.509 & 0.528 & 0.528 & 0.535 & \textbf{0.541} & 0.527 & 0.509 & 0.484 & 0.445 \\
MRR, $J{=}1$ & 0.277 & 0.278 & \textbf{0.279} & 0.278 & 0.274 & 0.267 & 0.260 & 0.245 & 0.233 \\
MRR, $J{=}2$ & 0.277 & \textbf{0.296} & 0.295 & 0.288 & 0.286 & 0.278 & 0.266 & 0.252 & 0.239 \\
MRR, $J{=}3$ & 0.283 & \textbf{0.300} & 0.299 & 0.295 & 0.292 & 0.280 & 0.270 & 0.252 & 0.241 \\
\midrule
Rerank share, $J{=}2$ & 0.00 & 0.27 & 0.32 & 0.43 & 0.54 & 0.81 & 1.08 & 1.62 & 2.17 \\
\bottomrule
\end{tabular}
\caption{[DEV] Rerank-weight sensitivity after range normalization on development data. Recall@10 and MRR exhibit a broad optimum across weight values and ensemble sizes, from which we select the deployed weight.}
\label{tab:beta}
\end{table*}

\subsection{Label-Coverage Diagnostics}
\label{app:coverage}

For query $q$ and candidate $c$ with canonical label $\lambda_c$,
% \begin{equation}
% \footnotesize
% \mathrm{cov}_{\lambda}(q,c)=\frac{|\,\mathrm{tok}(q)\cap\mathrm{tok}(\lambda_c)\,|}{|\,\mathrm{tok}(\lambda_c)\,|},
% \mathrm{cov}_{q}(q,c)=\frac{|\,\mathrm{tok}(q)\cap\mathrm{tok}(\lambda_c)\,|}{|\,\mathrm{tok}(q)\,|},
% \label{eq:coverage}
% \end{equation}
{\footnotesize
\begin{align}
\mathrm{cov}_{\lambda}(q,c) &= \frac{|\,\mathrm{tok}(q)\cap\mathrm{tok}(\lambda_c)\,|}{|\,\mathrm{tok}(\lambda_c)\,|}, \label{eq:coverage}\\
\mathrm{cov}_{q}(q,c) &= \frac{|\,\mathrm{tok}(q)\cap\mathrm{tok}(\lambda_c)\,|}{|\,\mathrm{tok}(q)\,|}, 
\end{align}
}

and the retriever of Section~\ref{sec:fintag} scores, under a fixed coverage weight $w_{\mathrm{cov}}$,
\begin{equation}
s(q,c)=\widehat{\mathrm{bm25}}(q,c)+w_{\mathrm{cov}}\bigl(\mathrm{cov}_{\lambda}(q,c)+\mathrm{cov}_{q}(q,c)\bigr),
\label{eq:retrieval_score}
\end{equation}

where $\widehat{\mathrm{bm25}}$ is the BM25 score range-normalized over the candidates scored for this query, so all three terms are on $[0,1]$ and $w_{\mathrm{cov}}$ is interpretable. The two coverage terms are not redundant: dividing by the label length rewards a candidate whose label is \emph{entirely} covered by the query, which rescues short generic concepts, while dividing by the query length rewards a candidate that accounts for \emph{more of what the query asked}, which separates candidates whose labels are equally short.

The term is a shared index component rather than an FHS contribution, so we retain only the diagnostics needed to establish its effect, its interaction with label-form rendering, and its mechanism (Table~\ref{tab:coveragegain}, Panels A and B).

\begin{table*}[!htbp]
\centering
\scriptsize
\begin{tabular}{@{}llcc@{\qquad}lrrr@{}}
\toprule
\multicolumn{4}{c}{\textbf{Panel A: Effect and interaction}} & \multicolumn{4}{c}{\textbf{Panel B: R@10 by gold-label length}} \\
\cmidrule(lr){1-4}\cmidrule(l){5-8}
\textbf{Analysis} & \textbf{Query/reference} & \textbf{$\Delta$R@10 [95\% CI]} & \textbf{$\Delta$MRR [95\% CI]} & \textbf{Length} & \textbf{$n$} & \textbf{Off} & \textbf{On} \\
\midrule
Gain & Raw context & $+0.231$ [0.090, 0.379] & $+0.135$ [0.049, 0.247] & 1 token & 44 & 0.000 & 0.568 \\
Gain & Free-text & $+0.175$ [0.085, 0.265] & $+0.088$ [0.038, 0.137] & 2 tokens & 62 & 0.000 & 0.258 \\
Gain & Structured label & $+0.219$ [0.134, 0.297] & $+0.137$ [0.085, 0.195] & 3--4 tokens & 275 & 0.058 & 0.295 \\
Interaction & Raw context & $-0.012$ [$-0.149$, 0.121] & $+0.001$ [$-0.100$, 0.089] & 5+ tokens & 280 & 0.189 & 0.304 \\
Interaction & Free-text & $+0.044$ [$-0.002$, 0.087] & $+0.049$ [0.013, 0.087] & & & & \\
\bottomrule
\end{tabular}
\caption{[DEV] Label-coverage diagnostics on the development sample: the coverage term's gain, its interaction with label-form rendering (Panel A, the tabular subset, $n{=}566$), and its effect by gold-label length (Panel B, all $661$ development facts). The coverage term is a shared index property applied to all methods, not an FHS-specific gain.}
\label{tab:coveragegain}
\label{tab:coverageinteraction}
\label{tab:coveragelength}
\end{table*}

Raw context receives the largest Recall@10 gain, while its MRR gain is comparable to that of structured label-form grounding. The interaction between the coverage term and label-form rendering is null against raw evidence. The gain also concentrates sharply on short generic labels: one- and two-token concepts are never retrieved within the top ten without the term, while concepts with five or more tokens gain substantially less than shorter concepts (Table~\ref{tab:coveragelength}, Panel B). These results support a length-correction rather than a semantic-alignment interpretation.

\section{Test-Split Diagnostics}
\label{app:test_dia}

\subsection{Retrieval-Readiness Diagnostics}
\label{app:probe}

These diagnostics support the paper's two headline characterizations of the retrieval-readiness gap, and are computed on the frozen test split after all design choices were fixed.

\paragraph{The gap is interpretive rather than index-side.}
When queried with the gold concept's own canonical label and definition, the retriever returns the target within the top ten for every test fact, at MRR $0.972$ (Table~\ref{tab:probe}). This is the empirical analogue of $g^*$ in Eq.~\ref{eq:gap}: the target is retrievable once the query expresses the intended concept.

\paragraph{The gap is primarily a precision gap.}
Raw context reaches the gold concept far more often at deep cutoffs than at shallow ones, whereas structured grounding raises shallow recall. The paired structured-versus-raw contrast at Recall@10 excludes zero for the definition form, $+0.113$ $[0.051, 0.178]$. At depth, grounding does not buy anything and may cost: the paired free-text-versus-raw contrast at Recall@50 spans zero on the negative side, $-0.066$ $[-0.182, 0.052]$. Grounding therefore mainly moves the target upward within an already reachable region rather than expanding reach. Table~\ref{tab:probe} reports the full query-form comparison. The rendering comparison between the definition and label forms is an ensemble-level design choice and is reported in Table~\ref{tab:ablation}.

A secondary diagnostic queries with the canonical \emph{label alone}, omitting the definition. Six concepts then fail to retrieve themselves at rank 1: \texttt{Assets}, \texttt{Liabilities}, \texttt{Revenues}, \texttt{Goodwill}, \texttt{Depreciation}, and \texttt{RegulatoryAssetsCurrent}. Longer compound concepts containing the queried term outrank the exact match, motivating the label-side normalization in Eq.~\eqref{eq:coverage}. The six failing concepts are listed here rather than tabulated; Table~\ref{tab:probe} reports the main probe.
% FHS

\subsection{Candidate-Level Verifier: Window, Judged Dimensions, and Sensitivity}
\label{app:window}

\paragraph{How the window is filled.} Section~\ref{sec:llmverifier} keeps the best-ranked candidate of each distinct category profile found in the top $60$. If that depth yields fewer than $K_v$ profiles, the remaining slots fall back to rank order, so $|\mathcal B_j|=K_v$ on every fact. As deployed, $\widetilde S$ does not depend on $j$, so all windows coincide, $\mathcal B_j=\mathcal B$, and a candidate is judged under every hypothesis or under none; we keep the indexed form because the machinery admits hypothesis-specific windows, and the per-arm windows of Table~\ref{tab:ablation} use them.

% \paragraph{Why every dimension is judged.} Restricting the verifier to the dimensions a candidate's surface text most directly exposes requires deciding, per inventory, which dimensions those are, and that decision does not transfer: in a clinical inventory laterality and encounter type are written into the code's own description, whereas in US-GAAP they are carried by structural metadata. Asking about the full schema needs no such judgement, and the verifier may abstain on a dimension the candidate does not settle -- the same convention that handles a dimension the hypothesis left unresolved. Verdicts are read only relative to the window: a candidate outside the window receives no candidate-specific
% verifier signal and is assigned the window-mean support, rather than
% an inferred contradiction. Scoring abstentions as non-support instead of leaving them out of the denominator moves the retrieval stage by at most $0.003$ on any metric, and not consistently in one direction: Recall@1 $+0.003$, Recall@10 $-0.003$, MRR $+0.003$, Recall@50 unchanged. The convention is therefore not load-bearing.

\paragraph{Judged dimensions and scoring conventions.}
The verifier evaluates all six dimensions because which attributes are
explicitly represented in a candidate varies across inventories. For
example, laterality and encounter type appear directly in many clinical
code descriptions, whereas analogous distinctions in US-GAAP may instead
be encoded in structural metadata. The verifier can abstain when either
the hypothesis or candidate does not provide enough information for a
dimension.

A candidate outside the verifier window receives the mean support within
that window rather than zero. Assigning zero would systematically favor
window membership and would therefore conflate the verifier's judgments
with the procedure used to construct its window. Within the window,
abstentions are excluded from both the numerator and denominator of the
support rate rather than being treated as negative verdicts. Under the
alternative convention that counts abstentions as non-support, retrieval
performance changes by at most 0.003: Recall@1 changes by $+0.003$,
Recall@10 by $-0.003$, MRR by $+0.003$, and Recall@50 is unchanged.
The reported results are therefore insensitive to this convention.

The candidate-level verifier evaluates only a short local window, so its cost and opportunity to change the ranking depend on $K_v$. We retain $K_v{=}10$ as the default operating point and evaluate smaller and larger windows without changing the generated hypotheses, retrieved pool, prompt, or downstream listwise selector. There is no separate verifier weight: the per-dimension verdicts are averaged into the single support value $\overline v(c)$, which enters the ranking under the rerank weight $\beta$ (Eq.~\ref{eq:hybrid_score}). Each window is a distinct generation run with that many candidates in the verifier's input, not a truncation of a larger window's judgements. Table~\ref{tab:llm_window_sensitivity} reports the sweep.

The two directions are not symmetric. Halving the window to $K_v{=}5$ costs Recall@10 ($-0.014$ $[-0.022,-0.008]$) and MRR ($-0.009$ $[-0.015,-0.004]$), so the deployed window is not larger than it needs to be. Doubling it to $K_v{=}20$ buys Recall@10 ($+0.014$ $[0.005, 0.024]$) but not head quality: MRR does not separate from the deployed window ($+0.002$ $[-0.004, 0.008]$) and neither does Recall@1 ($-0.003$ $[-0.010, 0.006]$). Judging more candidates therefore pulls additional gold concepts into the top ten while adding enough supported distractors above them to leave rank one no better. Recall@200 is identical at every window, as it must be: the verifier reorders the pool and cannot add to it. All intervals are the paired context-clustered bootstrap of Appendix~\ref{app:devsample} against the deployed $K_v{=}10$ arm.

\begin{table}[t]
\centering
\small
\begin{tabular}{@{}lccccc@{}}
\toprule
\textbf{Window $K_v$} & \textbf{R@1} & \textbf{R@10} & \textbf{R@50} & \textbf{R@200} & \textbf{MRR} \\
\midrule
5  & 0.179 & 0.383 & 0.543 & 0.705 & 0.248 \\
10 & \textbf{0.185} & \textbf{0.397} & \textbf{0.543} & \textbf{0.705} & \textbf{0.257} \\
20 & 0.182 & 0.411 & 0.546 & 0.705 & 0.259 \\
\bottomrule
\end{tabular}
\caption{Sensitivity to the candidate-level verifier window $K_v$ at the retrieval stage, pooled over the frozen test split ($n{=}2{,}509$). Bold marks the deployed configuration, $K_v{=}10$, whose retrieval columns reproduce Table~\ref{tab:ablation}'s FHS row to every printed digit. A wider window keeps lifting Recall@10 but not the head of the ranking. Recall@200 is identical across $K_v$ rows.}
\label{tab:llm_window_sensitivity}
\end{table}

\subsection{Candidate-Level Verifier: Behavior}
\label{app:verifier}
Section~\ref{sec:llmverifier} uses the LLM to compare candidates against a
hypothesis rather than to judge the hypothesis on its own.
Table~\ref{tab:verifierbridge} is why: Panel A measures the first ability and
Panel B the second, over the same verifier calls.
 
Two entries also motivate design choices elsewhere. The gold concept is inside
the assessed window on only 35.7\% of calls, which is what the profile-based
window construction of Section~\ref{sec:llmverifier} is for: filled by rank
alone, the window is dominated by lexical near-duplicates of one reading rather
than by competing readings. And the 25{,}699 dimension observations over 5{,}018
calls average 5.12 of the six dimensions, so the generator leaves roughly
one dimension in seven unresolved rather than guessing it
(Section~\ref{sec:generators}).
 
\begin{table}[!htbp]
\centering
\small
\begin{tabular}{@{}lr@{}}
\toprule
\multicolumn{2}{@{}l}{\emph{Panel A: candidate-level discrimination}} \\
LLM calls & 5{,}018 \\
Calls with gold in assessed window & 1{,}790 (35.7\%) \\
Support rate, gold candidate & 0.916 \\
Support rate, distractors & 0.543 \\
Gold $-$ distractor gap & $+0.373$ \\
Mean per-call gap & $+0.358$ \\
Calls favouring gold & 84.9\% \\
\midrule
\multicolumn{2}{@{}l}{\emph{Panel B: hypothesis-level calibration}} \\
Dimension observations & 25{,}699 \\
Hypothesis-wrong rate, all observations & 0.671 \\
LLM non-abstention rate & 0.793 \\
Hypothesis-wrong rate, judged observations & 0.604 \\
Mean support $\mid$ hypothesis wrong & 0.585 \\
Mean support $\mid$ hypothesis right & 0.597 \\
Difference & $-0.013$ \\
AUROC, support score vs.\ hypothesis wrong & 0.510 \\
\bottomrule
\end{tabular}
\caption{Candidate-level behavior of the LLM verifier on the frozen test split.
Panel A is within-window: how strongly the verifier separates the gold concept
from the distractors it sees alongside it. Panel B is across hypotheses: how
well the same support score predicts whether the hypothesis it was given is
itself wrong. The two abilities come apart. The verifier separates candidates
from one another by $+0.373$ and favours the gold candidate on 84.9\% of calls,
yet its score carries almost no signal about the hypothesis, at AUROC 0.510
against a 0.604 base rate. This is why the score enters only as a reranking
term over already-retrieved candidates (Eq.~\ref{eq:hybrid_score}) and never
selects among hypotheses.}
\label{tab:verifierbridge}
\end{table}
 
The verifier and the shared listwise selector could in principle be redundant:
both reorder the same candidates. Table~\ref{tab:verifier_reranker_interaction}
turns each on and off independently to separate them.

  % Off & Off & 0.205 & 0.124 & \num{2.3e-3} \\
  % On  & Off & 0.257 & 0.185 & \num{5.6e-3} \\
  % Off & On  & 0.319 & 0.240 & \num{1.8e-3} \\
  % On  & On  & \textbf{0.333} & \textbf{0.255} & \textbf{\num{2.8e-3}} \\
 
\begin{table}[!htbp]
\centering
\small
\begin{tabular}{@{}llccc@{}}
\toprule
\textbf{Verifier} & \textbf{Selector} & \textbf{MRR} & \textbf{Top-1} & \textbf{std} \\
\midrule
Off & Off & 0.205 & 0.124 & 2.3e-3 \\
On  & Off & 0.257 & 0.185 & 5.6e-3 \\
Off & On  & 0.319 & 0.240 & 7.5e-4 \\
On  & On  & \textbf{0.333} & \textbf{0.255} & 1.5e-3 \\
\bottomrule
\end{tabular}
\caption{Interaction between the candidate-level verifier and the shared
downstream listwise selector. Both columns are measured at whichever stage the
row ends, so unlike every other table the MRR column here is measured
\emph{after} the selector in the rows where it is on; the \textsf{On/Off} row
therefore reproduces Table~\ref{tab:ablation}'s FHS MRR and Recall@1, and the
\textsf{On/On} row its final accuracy. \textbf{Top-1} is Recall@1 in the
selector-off rows and Acc.\ in the selector-on rows, and std is over three runs
of the Top-1 column. The two stages are complementary rather than redundant:
the selector recovers most of the verifier's retrieval-stage gain on its own,
0.124 to 0.240, and the verifier still adds 0.015 of accuracy on top of it.}
\label{tab:verifier_reranker_interaction}
\end{table}
 
\subsection{Retriever Robustness Results}
\label{app:retriever_robustness}

The paper runs BM25 throughout, so a fair question is whether a stronger first-stage retriever would have changed the conclusion. Table~\ref{tab:retriever_robustness} repeats the comparison under dense and hybrid retrieval with everything else held at the deployed configuration, and reports each row's difference against the same method under BM25 with a paired interval. What this can settle is: two query-formation methods over one taxonomy index is enough to ask whether the FHS-versus-one-pass comparison survives a change of retriever. The retriever does change the candidate pool: Recall@200 moves by up to $0.031$ between the three. So the question is whether that propagates into the comparison: It does not. For one-pass grounding, dense retrieval gains $+0.013$ MRR $[+0.001,+0.026]$ while hybrid loses $-0.007$ $[-0.018,+0.004]$; for FHS the corresponding figures are $+0.008$ $[-0.005,+0.021]$ under dense and $-0.012$ $[-0.031,+0.006]$ under hybrid. Read as bounds rather than as null results, no swap admits an effect larger than $0.031$ MRR on either method, against a $+0.087$ MRR margin between the two methods under BM25, substantially larger than the observed effect of changing the
retriever. The same holds after the shared selector, where the largest interval bound reaches only $0.028$ accuracy. So within this comparison the ordering and its size come from how the query is formed rather than from which index answers it, and FHS keeps its margin under all three retrievers: $+0.087$, $+0.082$ and $+0.082$ MRR, and $+0.056$, $+0.057$ and $+0.049$ accuracy, over one-pass grounding under BM25, dense and hybrid respectively. Since neither swap closes the gap or reverses it, the remaining experiments use BM25, which is also the cheapest and the most reproducible of the three.

\begin{table}[t]
\centering
\setlength{\tabcolsep}{3pt}
\small
\resizebox{\columnwidth}{!}{%
\begin{tabular}{@{}llcccccccc@{}}
\toprule
& & \multicolumn{6}{c}{\textbf{Retrieval}} & \multicolumn{2}{c}{\textbf{Final}} \\
\cmidrule(lr){3-8}\cmidrule(lr){9-10}
\textbf{Retr.} & \textbf{Method} & \textbf{R@1} & \textbf{R@10} & \textbf{R@50} & \textbf{R@200} & \textbf{MRR} & $\Delta$\textbf{MRR} {\scriptsize[95\% CI]} & \textbf{Acc.} & $\Delta$\textbf{Acc.} {\scriptsize[95\% CI]} \\
\midrule
BM25 & One-pass & 0.106 & 0.304 & 0.496 & 0.734 & 0.171 & --- & 0.199 & --- \\
BM25 & FHS & 0.185 & 0.397 & 0.543 & 0.705 & 0.257 & --- & 0.255 & --- \\
\midrule
Dense & One-pass & 0.122 & 0.307 & 0.536 & 0.725 & 0.183 & {\scriptsize $+0.013$ [$+0.001,+0.026$]} & 0.198 & {\scriptsize $-0.001$ [$-0.016,+0.013$]} \\
Dense & FHS & 0.194 & 0.399 & 0.564 & 0.728 & 0.265 & {\scriptsize $+0.008$ [$-0.005,+0.021$]} & 0.254 & {\scriptsize $-0.001$ [$-0.019,+0.020$]} \\
\midrule
Hybrid & One-pass & 0.098 & 0.306 & 0.528 & 0.708 & 0.164 & {\scriptsize $-0.007$ [$-0.018,+0.004$]} & 0.194 & {\scriptsize $-0.005$ [$-0.019,+0.008$]} \\
Hybrid & FHS & 0.171 & 0.380 & 0.532 & 0.697 & 0.245 & {\scriptsize $-0.012$ [$-0.031,+0.006$]} & 0.243 & {\scriptsize $-0.012$ [$-0.028,+0.006$]} \\
\bottomrule
\end{tabular}}
\caption{Retriever robustness for one-pass grounding and FHS under sparse, dense, and hybrid retrieval. Every row runs the configuration of Table~\ref{tab:main_results} -- the label-coverage term on for both methods, and FHS scored by the candidate-level verifier at $K_v{=}10$ over its own fused window -- so the retriever is the only thing that differs, and the two BM25 rows are the runs Table~\ref{tab:main_results} reports. The dense and hybrid rows replay each method's logged hypotheses or logged query against the other retriever and then rerun the deployed scoring end to end; replayed under BM25 that procedure reproduces the deployed retrieval stage to within $10^{-6}$, which is what licenses reading the dense and hybrid rows against the BM25 ones. All columns except Acc.\ are measured before the shared listwise selector, per Section~\ref{sec:settings}. The $\Delta$ columns give each row's difference against the same method under BM25, with a 95\% interval from a paired bootstrap over source contexts (2{,}000 resamples), the estimator used throughout the paper; the same context resample is applied to both sides, since every run scores the same 2{,}509 facts.}
\label{tab:retriever_robustness}
\end{table}

\subsection{Results by Evidence Modality}
\label{app:modality}

  \begin{table*}[h]
  \centering
  \small
  \begin{tabular}{@{}lcccccccc@{}}
  \toprule
  & \multicolumn{4}{c}{\textbf{Table} } & \multicolumn{4}{c}{\textbf{Text}} \\
  \cmidrule(lr){2-5}\cmidrule(l){6-9}
  \textbf{Method} & \textbf{R@1} & \textbf{R@50} & \textbf{MRR} & \textbf{Acc.} & \textbf{R@1} &
  \textbf{R@50} & \textbf{MRR} & \textbf{Acc.} \\
  \midrule
  \multicolumn{9}{@{}l}{\emph{Single-pass and parallel}} \\
  Direct retrieval & 0.037 & 0.560 & 0.106 & 0.117 & 0.101 & 0.583 & 0.187 & 0.339 \\
  One-pass grounding, free-text & 0.101 & 0.484 & 0.163 & 0.185 & 0.167 & 0.655 & 0.277 & \textbf{0.393}
  \\
  One-pass grounding, structured & 0.121 & 0.551 & 0.200 & 0.218 & 0.131 & 0.601 & 0.237 & 0.339 \\
  Parallel sampling, stochastic ($J{=}2$) & 0.113 & 0.582 & 0.191 & 0.220 & 0.149 & 0.595 & 0.237 &
  0.345 \\
  Decomposed retrieval & 0.103 & 0.472 & 0.166 & 0.206 & 0.089 & 0.566 & 0.168 & 0.310 \\
  \textbf{FHS (full)} & \textbf{0.183} & 0.540 & \textbf{0.255} & \textbf{0.246} & \textbf{0.208} &
  0.589 & \textbf{0.286} & 0.381 \\
  \midrule
  \multicolumn{9}{@{}l}{\emph{Iterative and learned}} \\
  Intrinsic self-refinement & 0.126 & \textbf{0.617} & 0.214 & \textbf{0.224} & 0.131 & \textbf{0.661} &
  0.248 & 0.369 \\
  Retrieval-feedback refinement & \textbf{0.132} & \textbf{0.653} & \textbf{0.228} & 0.207 &
  \textbf{0.280} & \textbf{0.756} & \textbf{0.370} & \textbf{0.446} \\
  FHS-Seq & 0.112 & 0.538 & 0.192 & 0.217 & 0.107 & 0.625 & 0.204 & 0.345 \\
  \bottomrule
  \end{tabular}
  \caption{Results by evidence modality. Bold marks the two best values in each
  column. The two halves differ sharply in size: $2{,}341$ tabular facts against
  $168$ text ones, so a single text fact moves a text column by $0.6$ points, and
  the text half should be read as indicative rather than decisive. Column
  conventions follow Section~\ref{sec:settings}.}
  \label{tab:evidence_type}
  \end{table*}

  Both modalities are the same task, Financial Tagging, but they are not equally hard, and the
  difference is instructive. Serialized directly as a query, a text fact already
  reaches accuracy $0.339$ against $0.117$ for a tabular one: a narrative sentence
  names its concept far more often than a table cell does, because the cell's
  meaning is distributed over row and column headers that the value itself does
  not carry. Grounding is therefore worth much more on tables. FHS lifts tabular
  accuracy from $0.117$ to $0.246$, a gain of $+0.129$, against $+0.042$ on text.
  The interpretive gap is substantially more pronounced for tabular evidence.

  The one row that reverses the paper's ordering is retrieval-feedback
  refinement, which on text leads every column. Read with the same paired
  context-clustered bootstrap used throughout ($2{,}000$ resamples, resampling
  contexts rather than facts), the reversal is narrower than it looks. Its margin
  over FHS on text is significant for Recall@50 ($+0.167$ $[+0.090,+0.253]$) and
  MRR ($+0.084$ $[+0.009,+0.160]$), but not for Recall@1 ($+0.071$
  $[-0.017,+0.163]$) or accuracy ($+0.066$ $[-0.013,+0.148]$). On tables, where
  the sample is fourteen times larger, the same comparison separates cleanly and
  in both directions at once: the iterative baseline is ahead at Recall@50
  ($+0.113$ $[+0.057,+0.168]$) and behind at Recall@1 ($-0.051$
  $[-0.086,-0.021]$) and accuracy ($-0.039$ $[-0.064,-0.016]$).

  Taken together the two halves say the same thing rather than opposite things.
  Feeding retrieved candidates back into the next query is an effective way to
  \emph{reach} the gold concept: it wins Recall@50 on both modalities, and by
  more on text, but it does not consistently convert its coverage advantage into superior
top-1 ranking or final accuracy. And wherever the
  sample is large enough to resolve the head, it loses there. This is the
  distinction of Section~\ref{sec:precision} appearing within a single baseline:
  reach and position are separate quantities, and a method can buy one without
  the other. It also bounds what the text half licenses. With $168$ facts and
  $81$ contexts, the text column separates the two methods on reach and leaves
  the head unresolved; we therefore do not read it as evidence that iterative
  refinement beats FHS on narrative evidence, only that its coverage advantage
  survives the modality change.

\subsection{Computational Cost}
\label{subsec: cost}

Table~\ref{tab:efficiency} reports the average number of model calls per fact (LLM calls exclude the shared final listwise selector), together with end-to-end wall-clock time on the
CodiEsp set. The LLM call counts measure computational work per
fact, whereas wall-clock time is measured per full test-set run on a single
NVIDIA B200 GPU. Our current FHS implementation executes hypothesis generation
and verification serially, so the reported FHS wall time is not the optimal latency
that would be obtained from a fully parallel implementation.

% Non-integer
% retrieval counts arise because FHS may issue different numbers of queries for
% different facts and the table reports the average over the test set.

% \begin{table}[t]
% \centering
% \small
% \begin{tabular}{@{}lcrrc@{}}
% \toprule
% \textbf{Method} & \textbf{$J{/}B$} & \textbf{LLM} & \textbf{Ret.} & \textbf{Wall time (h)} \\
% \midrule
% Direct retr. & --- & 0.0 & 1.0 & $1.18 \pm 0.01$ \\
% One-pass, free-text & 1 & 1.0 & 1.0 & $1.20 \pm 0.02$ \\
% One-pass, structured & 1 & 1.0 & 1.8 & $2.42 \pm 0.04$ \\
% Parallel, stochastic & 2 & 2.0 & 2.0 & $2.03 \pm 0.05$ \\
% Decomposed & 4 & 1.0 & 4.0 & $2.74 \pm 0.06$ \\
% FHS (full) & 2 & 4.0 & 4.0 & $5.31 \pm 1.71$ \\
% \midrule
% \multicolumn{5}{@{}l}{\emph{Iterative and learned}} \\
% Intrinsic refine. & 4 & 4.0 & 4.0 & $4.86 \pm 0.01$ \\
% Feedback refine. & 4 & 4.0 & 4.0 & $5.56 \pm 0.17$ \\
% % FHS-Seq & 4 & 13.3 & 6.7 & --\\
% \bottomrule
% \end{tabular}
% \caption{Average inference cost per fact and wall-clock time on the CodiEsp test split. Wall time is
% reported as mean $\pm$ standard deviation over completed Slurm runs. J denotes the number of parallel hypotheses, and B denotes the maximum number of refinement rounds.}
% \label{tab:efficiency}
% \end{table}
\begin{table}[t]
\centering
\small
\begin{tabular}{@{}lcrc@{}}
\toprule
\textbf{Method} & \textbf{$J{/}B$} & \textbf{LLM} & \textbf{Wall time (h)} \\
\midrule
Direct retr. & --- & 0.0 & $1.18 \pm 0.01$ \\
One-pass, free-text & 1 & 1.0 & $1.20 \pm 0.02$ \\
One-pass, structured & 1 & 1.0 & $2.42 \pm 0.04$ \\
Parallel, stochastic & 2 & 2.0 & $2.03 \pm 0.05$ \\
Decomposed & -- & 1.0 & $2.74 \pm 0.06$ \\
FHS (full) & 2 & 4.0 & $5.31 \pm 1.71$ \\
\midrule
\multicolumn{4}{@{}l}{\emph{Iterative and learned}} \\
Intrinsic refine. & 4 & 4.0 & $4.86 \pm 0.01$ \\
Feedback refine. & 4 & 4.0 & $5.56 \pm 0.17$ \\
% FHS-Seq & 4 & 13.3 & --\\
\bottomrule
\end{tabular}
\caption{Average inference cost per fact and wall-clock time on the CodiEsp test split. Wall time is
reported as mean $\pm$ standard deviation over completed Slurm runs. J denotes the number of parallel hypotheses, and B denotes the maximum number of refinement rounds.}
\label{tab:efficiency}
\end{table}

% Table~\ref{tab:efficiency} reports the average numbers of model calls and retrieval operations per fact. The generator and verifier calls can be parallelized across hypotheses for FHS, so these counts measure computational work rather than end-to-end latency. We report call counts instead of wall-clock time because runtime on a shared cluster is strongly affected by scheduling and system load. Non-integer retrieval counts arise because FHS may issue different numbers of queries for different facts and the table reports the average over the test set.

% \begin{table}[t]
% \centering
% \small
% \begin{tabular}{@{}lcrr@{}}
% \toprule
% \textbf{Method} & \textbf{$J{/}B$} & \textbf{LLM} & \textbf{Ret.} \\
% \midrule
% Direct retr. & --- & 0.0 & 1.0 \\
% One-pass, free-text & 1 & 1.0 & 1.0 \\
% One-pass, structured & 1 & 1.0 & 1.8 \\
% Parallel, stochastic & 2 & 2.0 & 2.0 \\
% Decomposed & 4 & 1.0 & 4.0 \\
% FHS (full) & 2 & 4.0 & 3.9 \\
% \midrule
% \multicolumn{4}{@{}l}{\emph{Iterative and learned}} \\
% Intrinsic refine. & 4 & 4.0 & 4.0 \\
% Feedback refine. & 4 & 4.0 & 4.0 \\
% FHS-Seq & 4 & 15.5 & 8.3 \\
% \bottomrule
% \end{tabular}
% \caption{Average inference cost per fact on the Financial Tagging test split }
% \label{tab:efficiency}
% \end{table}

%% =====================================================================

%% =====================================================================

%% =====================================================================
\section{Sequential Refinement: A Negative Result}

\subsection{Sequential Refinement: Full Analysis}
\label{app:seq}
This section expands the negative control of Section~\ref{sec:negative} on the full test split. FHS-Seq differs from FHS only in control flow: one parallel fan versus a sequential loop over the same generator, renderer, retriever, aggregator, and candidate-level verifier. Table~\ref{tab:seq_outcome} carries the numbers.

\paragraph{Iteration does help over a weak single pass.}
This must be stated first, because it is the comparison the prior literature makes and we reproduce it. On the test split, free-form iterative baselines exceed one-pass grounding (Table~\ref{tab:seq_results}). Repeated revision of a single weak grounding is beneficial in this task.

\paragraph{It does not help over a strong parallel first round.}
Across all 2{,}509 test facts with a four-round budget, round one follows the same procedure as FHS's parallel round, at Recall@50 $0.550$, and the full episode ends at $0.544$ (Table~\ref{tab:seq_outcome}). The difference is $-0.006$ $[-0.016, 0.002]$ under the paired context-clustered bootstrap, nominally negative and not distinguishable from zero, even though the later rounds are not idle: they replace $13.8\%$ of the top-50 pool and consume $3.3$ of the four rounds on average. The head is where the cost shows. The sequential arm ends below FHS on every column that measures rank rather than reach: $0.362$ against $0.397$ at Recall@10, $0.193$ against $0.257$ MRR, and $0.226$ against $0.255$ final accuracy (Table~\ref{tab:main_results}).

% \paragraph{Why iteration has no room here.}
% The parallel architecture already reaches 91\% of the best-single-hypothesis oracle on the test split: 0.397 against 0.437 Recall@10, with both arms carrying the same verifier over the same six dimensions (Table~\ref{tab:ablation}). The head is where the residual sits: 0.185 against 0.222 at Recall@1. Knowing which dimension is wrong locates the error but does not fix it, and the residual gap is too small for an imperfect correction rule to close. On this task single-pass generation is already near the achievable ceiling, so sequential refinement has little room to pay; we expect it to earn its cost where single-pass generation is far from that ceiling. Two secondary observations support the same reading. The reward signal is thin: on $0.298$ of facts the gold concept never enters the accumulated pool at all. And the loop spends its full four-round budget on $0.733$ of instances while still not improving on round one (Appendix~\ref{app:seqdiag}).

\paragraph{Oracle definition and headroom calculation.}
For each test instance, the oracle evaluates the same hypotheses generated by
FHS and selects the single hypothesis that gives the gold concept its highest
final rank. Each hypothesis uses the same rendering, retrieval, and
candidate-level verification procedure as the deployed method. The oracle
therefore does not introduce a new hypothesis or use a different scoring
pipeline. It provides an upper bound on what a perfect hypothesis-selection
rule could achieve over this fixed generated set, rather than an upper bound
on FHS in general. For a metric $m$, we report the remaining absolute headroom
as
\[
H_m = m_{\mathrm{oracle}} - m_{\mathrm{FHS}},
\]
and the fraction of the oracle performance attained by FHS as
\[
A_m = \frac{m_{\mathrm{FHS}}}{m_{\mathrm{oracle}}}.
\]

\paragraph{Headroom results.}
FHS reaches 0.185 Recall@1, 0.397 Recall@10, 0.543 Recall@50, 0.257 MRR,
and 0.255 final accuracy. Oracle selection over the same hypotheses reaches
0.222, 0.437, 0.592, 0.295, and 0.262, respectively. The corresponding
absolute headroom is therefore 0.037 at Recall@1, 0.040 at Recall@10, 0.049
at Recall@50, 0.038 in MRR, and 0.007 in final accuracy. Equivalently, FHS
attains 83.3\%, 90.8\%, 91.7\%, 87.1\%, and 97.3\% of the oracle values on
these metrics. Thus, some head-ranking headroom remains, but perfect selection
among the existing hypotheses would yield only a small improvement in final
accuracy. This result helps explain why imperfect sequential revisions do not
outperform the strong parallel round. Two additional diagnostics support this interpretation. For 29.8\% of test
facts, the gold concept never enters the accumulated candidate pool, so
hypothesis revision receives no direct signal about it. Moreover,
\mbox{FHS-Seq} consumes its full four-round budget on 73.3\% of instances
without improving over round one. Full round-level diagnostics are reported
in Appendix~\ref{app:seqdiag}.

\subsection{Sequential Control: Full-Test Diagnostics}
\label{app:seqdiag}

All results use the full 2{,}509-fact test split with an identical instance order and a four-round budget, and test whether later rounds improve the loop over its own parallel first round.

Table~\ref{tab:seq_outcome} carries the comparison: Recall@50 after round one against the full episode, and how much of the top-50 pool the later rounds replace.

\begin{table}[H]
\centering
\small
\resizebox{0.92\columnwidth}{!}{%
\begin{tabular}{@{}lrrrr@{}}
\toprule
& \multicolumn{4}{c}{\textbf{Round 1 vs. full episode}} \\
\cmidrule(l){2-5}
\textbf{Arm} & \textbf{Rd-1} & \textbf{Full} & \textbf{$\Delta$} & \textbf{Chg. \%} \\
\midrule
FHS-Seq & 0.550 & 0.544 & $-0.006$ & 13.8 \\
\bottomrule
\end{tabular}%
}
\caption{Sequential control on the full test split: Recall@50 for round one against the full episode, and the share of the top-50 pool the later rounds replace. Rd-1 follows the same procedure as FHS's parallel round, the control opens with it, and both columns are read off the same run at the same stage, before the candidate-level verifier. Rd-1 uses this run's own stochastic samples, so it differs from the three-run mean of $0.543$ reported for FHS in Table~\ref{tab:main_results} by sampling variation. The paired context-clustered interval on $\Delta$ is $[-0.016, 0.002]$.}
\label{tab:seq_outcome}
\end{table}

The added rounds replace $13.8\%$ of the top-50 pool and move Recall@50 by $-0.006$ $[-0.016, 0.002]$: they change the pool without improving it. The reward signal behind them is thin. On $0.298$ of facts the gold concept never enters the accumulated pool at all, so no amount of revision can promote it. The loop nonetheless spends its full four-round budget on $0.733$ of instances; the rest stop early only because no dimension is left unsupported, $0.174$ of them after a single round.

\section{Qualitative Case Studies}
\label{app:cases}

The tables in Section~\ref{sec:experiments} report average performance. This appendix instead follows four individual facts through the FHS pipeline. The first two are financial facts, one tabular and one narrative. The other two are CodiEsp diagnosis facts. We selected these cases to illustrate the mechanism, not to represent either domain or evidence type.

All quantities below come from the deployed pipeline. Here, $\widetilde S$ is the range-normalized fused score from Section~\ref{sec:aggregator}, and $\overline v$ is the average verifier support from Eq.~\ref{eq:llm_support}. The final score is $S_{\mathrm{final}}=\widetilde S+\beta\overline v$, with $\beta{=}0.6$ (Eq.~\ref{eq:hybrid_score}). The \emph{w/o verifier} column reports the ranking when the verifier term is removed. On the financial split, the verifier moves the gold concept to rank one for $156$ facts and away from rank one for $5$. In the deployed CodiEsp run, the corresponding counts are $221$ and $78$.

\subsection{A tabular fact: a share-based compensation rollforward}
\label{app:case_table}

The source is an award rollforward with two columns and five rows. The columns report share count and weighted-average grant-date fair value. The rows report the opening balance, grants, vesting, forfeitures, and the closing balance. The located fact is $1{,}229{,}202$ in the row \emph{Nonvested at December 31, 2023} and the column \emph{Total Number of RSUs}. Its datatype is \texttt{sharesItemType}.

The cell does not name its concept. Its context supports several concepts that the taxonomy treats separately. These include nonvested equity instruments other than options, nonvested options, shares available for grant, and the period change in nonvested instruments. These are different interpretations of the same cell. Section~\ref{sec:factorized} represents this type of ambiguity with factorized hypotheses. The two sampled hypotheses agree on the main reading but describe it differently (Table~\ref{tab:case_hyp_table}). The first uses \emph{Share-Based Compensation} as the \textsc{Family} and \emph{Nonvested Shares} as the \textsc{Role}. The second uses \emph{Equity} as the \textsc{Family} and \emph{Share-based compensation} as the \textsc{Role}. They also express \textsc{Temporal} differently. Both leave \textsc{Scope} unresolved. This variation arises from independent sampling rather than an instruction to produce different hypotheses.

\begin{table}[!htbp]
\centering
\small
\setlength{\tabcolsep}{3.5pt}
\begin{tabular}{@{}lll@{}}
\toprule
\textbf{Dimension} & \textbf{$h_1$} & \textbf{$h_2$} \\
\midrule
\textsc{Family}    & Share-Based Comp. & Equity \\
\textsc{Role}      & Nonvested Shares & Share-based comp. \\
\textsc{Event}     & Nonvested at & Nonvested at \\
                   & Dec.\ 31, 2023 & Dec.\ 31, 2023 \\
\textsc{Qualifier} & Total Number & Total Number \\
\textsc{Scope}     & $\bot$ & $\bot$ \\
\textsc{Temporal}  & point\_in\_time & December 31, 2023 \\
\bottomrule
\end{tabular}
\caption{The two factorized hypotheses sampled for the tabular fact of Appendix~\ref{app:case_table}. $\bot$ marks an unresolved dimension. Each is issued in both renderings of Section~\ref{sec:render}, so four rankings are fused.}
\label{tab:case_hyp_table}
\end{table}

\begin{table*}[!htbp]
\centering
\small
\begin{tabular}{@{}lccccc@{}}
\toprule
\textbf{Candidate concept} & \textbf{$\widetilde S$} & \textbf{$\overline v$} & \textbf{$S_{\mathrm{final}}$} & \textbf{w/o verifier} & \textbf{FHS} \\
\midrule
Nonvested number, other than options \;(\textbf{gold}) & 0.8819 & 1.0000 & 1.4819 & 7 & \textbf{1} \\
Nonvested number of shares, options & 0.9446 & 0.8750 & 1.4696 & 3 & 2 \\
Number of shares available for grant & 1.0000 & 0.2917 & 1.1750 & 1 & 3 \\
Stock issued in period, share-based comp.\ forfeited & 0.9385 & 0.3750 & 1.1635 & 4 & 4 \\
Other than options, period increase/decrease & 0.9534 & 0.3500 & 1.1634 & 2 & 5 \\
Nonvested options forfeited, number of shares & 0.9132 & 0.3850 & 1.1442 & 6 & 6 \\
Incremental common shares, forfeitable dividends & 0.9290 & 0.3333 & 1.1290 & 5 & 7 \\
\bottomrule
\end{tabular}
\caption{The head of the ranking for the tabular fact, before and after the candidate-level verifier. Concept names are shortened. The two that decide the case are \texttt{ShareBasedCompensation\-ArrangementByShareBasedPaymentAward\-EquityInstrumentsOtherThanOptions\-NonvestedNumber} (gold) and \texttt{SharebasedCompensation\-ArrangementBySharebasedPaymentAward\-OptionsNonvestedNumberOfShares}, both in the \texttt{us-gaap} namespace. The last candidate was never inside a verifier window and carries the window mean of Eq.~\ref{eq:llm_support} rather than a judged value.}
\label{tab:case_rank_table}
\end{table*}

Fusion alone does not resolve this fact. In the fused ranking, several concepts share the long prefix \texttt{ShareBasedCompensationArrangement\-ByShareBasedPaymentAward} but differ in their suffixes. The gold concept is ranked seventh (Table~\ref{tab:case_rank_table}). This case motivates the profile-diverse window in Section~\ref{sec:llmverifier}. The window covers distinct category profiles instead of taking only the top $K_v$ candidates by score. The verifier separates the leading candidates using the named dimensions. Under $h_1$, it supports the gold concept on \textsc{Family}, \textsc{Role}, \textsc{Event}, and \textsc{Temporal}. It abstains on the other two dimensions, so $v_1=4/4=1.0$. For the options concept, it gives the same verdicts except for \textsc{Role}, which yields $v_1=3/4=0.75$. Under $h_2$, both concepts receive $1.0$ support. Their average support scores are therefore $1.000$ and $0.875$. After weighting by $\beta$, the support gap is $0.075$. This exceeds the $0.063$ fused-score advantage of the options concept. The gold concept therefore moves to rank one. The decisive evidence comes from one \textsc{Role} verdict under one hypothesis. Dimension-level verdicts make this difference explicit and auditable. A single candidate-level score could produce the same ordering, but it would not show which semantic distinction caused it.

\subsection{A narrative fact: an amended securitization facility}
\label{app:case_text}

The second fact appears in narrative text. The passage states that a receivables securitization facility ``was amended in August 2024 to extend the maturity date $\ldots$ and increase the aggregate commitments from \$$370$ million to \$$400$ million.'' The located value is $370$. Its concept is the facility's maximum borrowing capacity before the amendment.

The sentence focuses on an amendment event, while the target concept describes capacity. Several baseline rewrites follow the event framing. Examples include ``increase in securitization facility commitments'' and ``increase in accounts receivable securitization facility commitments due to amendment.'' The gold concept then appears at ranks $44$, $172$, and $194$ for these methods (Table~\ref{tab:case_methods}).

\begin{table}[!htbp]
\centering
\small
\setlength{\tabcolsep}{3.5pt}
\begin{tabular}{@{}lll@{}}
\toprule
\textbf{Dimension} & \textbf{$h_1$} & \textbf{$h_2$} \\
\midrule
\textsc{Family}    & Liabilities & Liabilities \\
\textsc{Role}      & Debt Facility & Debt \\
                   & Commitments & \\
\textsc{Event}     & Amended securitization & Borrowing \\
                   & facility and increased & \\
                   & commitments & \\
\textsc{Qualifier} & $\bot$ & $\bot$ \\
\textsc{Scope}     & $\bot$ & $\bot$ \\
\textsc{Temporal}  & Point in Time & $\bot$ \\
\bottomrule
\end{tabular}
\caption{The two factorized hypotheses sampled for the narrative fact of Appendix~\ref{app:case_text}. On narrative evidence only the definition form is issued (Section~\ref{sec:render}), so two rankings are fused rather than four.}
\label{tab:case_hyp_text}
\end{table}

\begin{table*}[!htbp]
\centering
\small
\begin{tabular}{@{}lccccc@{}}
\toprule
\textbf{Candidate concept} & \textbf{$\widetilde S$} & \textbf{$\overline v$} & \textbf{$S_{\mathrm{final}}$} & \textbf{w/o verifier} & \textbf{FHS} \\
\midrule
Line of credit facility, maximum borrowing capacity \;(\textbf{gold}) & 0.8721 & 0.500 & 1.1721 & 2 & \textbf{1} \\
Line of credit facility, remaining borrowing capacity & 0.8105 & 0.500 & 1.1105 & 4 & 2 \\
Line of credit facility, current borrowing capacity & 0.7901 & 0.500 & 1.0901 & 6 & 3 \\
Debtor reorganization items, debtor-in-possession facility financing costs & 1.0000 & 0.000 & 1.0000 & 1 & 4 \\
Facility costs & 0.8680 & 0.000 & 0.8680 & 3 & 5 \\
Reorganization items & 0.7919 & 0.000 & 0.7919 & 5 & 6 \\
\bottomrule
\end{tabular}
\caption{The head of the ranking for the narrative fact, before and after the candidate-level verifier. The fused ranking interleaves the line-of-credit family with concepts that share only the token \emph{facility}; the verifier assigns the two groups $0.5$ and $0.0$ and separates them. The first three rows are \texttt{us-gaap:LineOfCreditFacility\{Maximum, Remaining, Current\}BorrowingCapacity}; the remaining names shorten the concepts' canonical labels.}
\label{tab:case_rank_text}
\end{table*}

FHS samples one hypothesis that follows the amendment framing and one that does not (Table~\ref{tab:case_hyp_text}). The first resolves \textsc{Event} as \emph{Amended securitization facility and increased commitments}. The second resolves only \textsc{Family} and \textsc{Role} as \emph{Liabilities} and \emph{Debt}. It leaves the other four dimensions unresolved. Under the first hypothesis, the verifier rejects the gold concept on \textsc{Event} and abstains on the other dimensions. This gives $v_1=0/1=0.0$. Under the second, it supports the gold concept on \textsc{Family} and \textsc{Role} and abstains on the rest. This gives $v_2=2/2=1.0$. The average support is $0.5$. Table~\ref{tab:case_rank_text} shows the effect of this support. The fused ranking mixes three line-of-credit capacity concepts with three concepts that mainly share the word \emph{facility}. The verifier gives the second group $\overline v=0.0$ under both hypotheses. The three capacity concepts each receive $0.5$. Adding $\beta\overline v$ moves the capacity family above the unrelated concepts. Within that family, the fused score places maximum capacity above remaining and current capacity. Thus, the verifier separates the families, while the fused score orders concepts within the selected family. Two design choices in Eq.~\ref{eq:llm_support} matter for this fact. If FHS used only the first hypothesis, the gold concept would receive the same $0.0$ support as the distractors. The second partial hypothesis preserves an alternative reading. Its support is also not diluted by unresolved dimensions. If abstentions counted as non-support, its two positive verdicts would be divided by six rather than two. The resulting average support for the gold concept would fall from $0.5$ to $0.167$. Appendix~\ref{app:window} evaluates this scoring convention across the full split.

\subsection{What the first-domain cases have in common}

\begin{table}[!htbp]
\centering
\small
\setlength{\tabcolsep}{4pt}
\begin{tabular}{@{}lcc@{}}
\toprule
& \textbf{Tabular} & \textbf{Narrative} \\
\textbf{Method} & \textbf{fact} & \textbf{fact} \\
\midrule
Direct retrieval & 29 & 20 \\
One-pass, free-text & 63 & 44 \\
One-pass, structured & 8$^{\checkmark}$ & 13$^{\checkmark}$ \\
Parallel, stochastic ($J{=}2$) & 96 & $>$200 \\
Decomposed & 23 & $>$200 \\
Intrinsic refinement & 17 & 194 \\
Feedback refinement & 81 & 172 \\
\midrule
FHS $-$ verifier & 7$^{\checkmark}$ & 2 \\
\textbf{FHS} & \textbf{1}$^{\checkmark}$ & \textbf{1}$^{\checkmark}$ \\
\bottomrule
\end{tabular}
\caption{Rank of the gold concept at the retrieval stage for the two facts, by method; $>$200 means the concept is absent from the $200$-candidate pool. A check mark marks the methods whose shared listwise selector then returned the gold concept as its top-1. Row names follow Table~\ref{tab:main_results}.}
\label{tab:case_methods}
\end{table}

Table~\ref{tab:case_methods} compares all methods on the two financial facts. For the tabular fact, every alternative retrieves the gold concept within the top 100, but none places it within the top five. The problem is therefore ranking rather than candidate coverage. For the narrative fact, the two iterative methods place the gold concept at ranks $172$ and $194$. Direct retrieval places it at rank $20$, while parallel sampling and decomposed retrieval miss it from the top 200. These cases show that additional retrieval does not necessarily help when the query follows the wrong interpretation.

The two financial cases fail in different ways. In the tabular case, the hypotheses capture the intended reading, but retrieval does not separate closely related labels. One \textsc{Role} verdict resolves the ambiguity. In the narrative case, one hypothesis follows the wrong event framing. The other preserves a broader reading, and the verifier rejects unrelated candidates. Both cases use the same mechanism: a partial assignment over named dimensions lets each verdict refer to a specific semantic property. This observation is consistent with the ablation in Table~\ref{tab:ablation}. Replacing factorized hypotheses with an equally sized free-text ensemble, without the verifier, reduces Recall@1 by $0.069$ ($0.037$ with the verifier restored).

\subsection{CodiEsp diagnosis cases}
\label{app:case_codiesp}

CodiEsp uses a different evidence type and taxonomy. Each fact is a relocated diagnosis mention from the English machine-translated version of CodiEsp. Spanish offsets and references are used only to locate the mention and preserve provenance. And each candidate is an ICD--10--CM diagnosis code. The verifier uses the six dimensions defined in Appendix~\ref{app:codiesp}. The experiments use the full exact-relocation test split with $w_{\mathrm{cov}}{=}1.0$. Tables~\ref{tab:case_codiesp_reflux_rank} and \ref{tab:case_codiesp_metastasis_rank} use the candidate sets from the deployed FHS run. The \emph{w/o verifier} column reorders each set using $\widetilde S$ alone. Table~\ref{tab:case_codiesp_methods} instead reports a separately run FHS $-$ verifier arm together with the other baselines.

\paragraph{Gastric reflux.}
The first CodiEsp fact is the mention \emph{gastric reflux}. It appears in a note about a patient with a history of gastrectomy and Billroth II reconstruction. The gold code is \texttt{K21.9}, gastro-esophageal reflux disease without esophagitis. The ambiguity is lexical. A query containing \emph{reflux} retrieves many vesicoureteral reflux codes, although the note places the condition in the gastric context. The two hypotheses state the intended reading directly. One sets \textsc{Event} to \emph{Gastric reflux}; the other uses \emph{gastroesophageal reflux}. Both assign the case to digestive diseases and leave \textsc{Qualifier} unresolved. Without the verifier, \texttt{N13.70}, vesicoureteral reflux, ranks first. The gold code ranks third (Table~\ref{tab:case_codiesp_reflux_rank}). The verifier gives full support to both gastro-esophageal reflux codes and zero support to the vesicoureteral reflux codes. It therefore moves the correct disease family above the urinary-tract codes. The fused score then places the gold code without esophagitis above its sibling with esophagitis.

\begin{table*}[!htbp]
\centering
\small
\begin{tabular}{@{}lccccc@{}}
\toprule
\textbf{Candidate code} & \textbf{$\widetilde S$} & \textbf{$\overline v$} & \textbf{$S_{\mathrm{final}}$} & \textbf{w/o verifier} & \textbf{FHS} \\
\midrule
\texttt{K21.9} Gastro-esophageal reflux disease, no esophagitis \;(\textbf{gold}) & 0.9833 & 1.0000 & 1.5833 & 3 & \textbf{1} \\
\texttt{K21.0} Gastro-esophageal reflux disease with esophagitis & 0.9579 & 1.0000 & 1.5579 & 5 & 2 \\
\texttt{K31.4} Gastric diverticulum & 0.9988 & 0.1667 & 1.0988 & 2 & 3 \\
\texttt{I86.4} Gastric varices & 0.9176 & 0.1667 & 1.0176 & 8 & 4 \\
\texttt{N13.70} Vesicoureteral-reflux, unspecified & 1.0000 & 0.0000 & 1.0000 & 1 & 5 \\
\texttt{N13.71} Vesicoureteral-reflux without reflux nephropathy & 0.9819 & 0.0000 & 0.9819 & 4 & 6 \\
\texttt{N13.739} Vesicoureteral-reflux with hydroureter, unspecified & 0.9473 & 0.0000 & 0.9473 & 6 & 7 \\
\bottomrule
\end{tabular}
\caption{CodiEsp case: \emph{gastric reflux}, gold \texttt{K21.9}. The no-verifier rank sorts the same FHS candidate set by $\widetilde S$ alone; FHS adds $\beta\overline v$ with $\beta{=}0.6$.}
\label{tab:case_codiesp_reflux_rank}
\end{table*}

\paragraph{Pulmonary metastasis.}
The second CodiEsp fact requires more clinical context. The note reports a prior right nephrectomy for renal carcinoma and a later right pneumonectomy for pulmonary metastasis. The gold code is \texttt{C78.01}, secondary malignant neoplasm of right lung. The word \emph{pulmonary} attracts codes for pulmonary hypertension, infection, and embolism. However, \emph{metastasis}, the earlier renal carcinoma, and the right pneumonectomy indicate a secondary malignant neoplasm of the right lung. Here the two hypotheses are almost identical. Both identify a malignant neoplasm that has metastasized to the right lung. Their queries differ in form: one is code-like, \texttt{C78.0}, while the other is phrase-like, \emph{pulmonary metastasis malignant right}. Fusion alone leaves the gold code at rank eight because candidates containing \emph{pulmonary} dominate the list. The verifier supports \texttt{C78.01} on all six dimensions under both hypotheses. It gives zero support to the non-neoplasm pulmonary candidates. The gold code therefore moves to rank one (Table~\ref{tab:case_codiesp_metastasis_rank}).

\begin{table*}[!htbp]
\centering
\small
\begin{tabular}{@{}lccccc@{}}
\toprule
\textbf{Candidate code} & \textbf{$\widetilde S$} & \textbf{$\overline v$} & \textbf{$S_{\mathrm{final}}$} & \textbf{w/o verifier} & \textbf{FHS} \\
\midrule
\texttt{C78.01} Secondary malignant neoplasm of right lung \;(\textbf{gold}) & 0.6802 & 1.0000 & 1.2802 & 8 & \textbf{1} \\
\texttt{I27.0} Primary pulmonary hypertension & 1.0000 & 0.0000 & 1.0000 & 1 & 2 \\
\texttt{B42.0} Pulmonary sporotrichosis & 0.8396 & 0.0000 & 0.8396 & 2 & 3 \\
\texttt{B46.0} Pulmonary mucormycosis & 0.8250 & 0.0000 & 0.8250 & 3 & 4 \\
\texttt{B45.0} Pulmonary cryptococcosis & 0.8001 & 0.0000 & 0.8001 & 4 & 5 \\
\texttt{B41.0} Pulmonary paracoccidioidomycosis & 0.7864 & 0.0000 & 0.7864 & 5 & 6 \\
\texttt{I26.09} Other pulmonary embolism with acute cor pulmonale & 0.7662 & 0.0000 & 0.7662 & 6 & 7 \\
\texttt{C79.9} Secondary malignant neoplasm of unspecified site & 0.6429 & 0.0000 & 0.6429 & 11 & 11 \\
\bottomrule
\end{tabular}
\caption{CodiEsp case: \emph{pulmonary metastasis}, gold \texttt{C78.01}. The no-verifier rank sorts the same FHS candidate set by $\widetilde S$ alone; FHS adds $\beta\overline v$ with $\beta{=}0.6$.}
\label{tab:case_codiesp_metastasis_rank}
\end{table*}

\begin{table}[!htbp]
\centering
\small
\setlength{\tabcolsep}{4pt}
\begin{tabular}{@{}lcc@{}}
\toprule
& \textbf{Gastric} & \textbf{Pulmonary} \\
\textbf{Method} & \textbf{reflux} & \textbf{metastasis} \\
\midrule
Direct retrieval & 66 & $>$200 \\
One-pass, free-text & 11 & 7 \\
One-pass, structured & \textbf{1}$^{\checkmark}$ & 27 \\
Parallel, stochastic ($J{=}2$) & 2 & \textbf{1}$^{\checkmark}$ \\
Decomposed & 7$^{\checkmark}$ & $>$200 \\
Intrinsic refinement & 13$^{\checkmark}$ & 2$^{\checkmark}$ \\
Feedback refinement & 2$^{\checkmark}$ & \textbf{1}$^{\checkmark}$ \\
\midrule
FHS $-$ verifier & 7$^{\checkmark}$ & 49$^{\checkmark}$ \\
\textbf{FHS} & \textbf{1}$^{\checkmark}$ & \textbf{1}$^{\checkmark}$ \\
\bottomrule
\end{tabular}
\caption{Rank of the gold ICD--10--CM code at the retrieval stage for the two CodiEsp cases; $>$200 means the code is absent from the $200$-candidate pool. A check mark marks the methods whose shared listwise selector returned the gold code as top-1.}
\label{tab:case_codiesp_methods}
\end{table}

Table~\ref{tab:case_codiesp_methods} compares the two cases across methods. For gastric reflux, several methods retrieve the correct family, but their final selectors differ in whether they recover the gold code. For pulmonary metastasis, some baselines rank the gold code highly. Direct and decomposed retrieval miss it from the top 200, while the separately run FHS $-$ verifier arm places it at rank $49$. In both cases, the verifier rejects candidates that conflict with the named dimensions. This is the same role it plays in the financial cases, even though the taxonomy, evidence, and source of ambiguity differ.

%% =====================================================================
\end{document}